\documentclass[10pt,twocolumn,letterpaper]{article}

\usepackage[pagenumbers]{cvpr} %
\usepackage{cvpr} %

\newcommand{\secondplace}[1]{\textcolor{blue}{\textbf{#1}}}

\definecolor{cvprblue}{rgb}{0.21,0.49,0.74}
\usepackage[pagebackref,breaklinks,colorlinks,allcolors=cvprblue]{hyperref}

\usepackage[ruled,linesnumbered]{algorithm2e}
\usepackage{mathtools}
\usepackage{multirow}
\usepackage{wrapfig}
\usepackage{xcolor}
\usepackage{svg}
\usepackage{float}
\usepackage{amssymb}
\usepackage{amsfonts}
\usepackage{url}
\usepackage{enumitem}
\usepackage{subcaption}
\usepackage[accsupp]{axessibility}

\makeatletter
\renewcommand\paragraph{\@startsection{paragraph}{4}{\z@}%
  {1.5ex \@plus 0.5ex \@minus 0.1ex}
  {-1em}
  {\normalfont\normalsize\bfseries}} 
\makeatother

\title{Denoising Functional Maps: Diffusion Models for Shape Correspondence}

\author{Aleksei Zhuravlev\textsuperscript{1} \quad\quad Zorah Lähner\textsuperscript{1, 2} \quad\quad Vladislav Golyanik\textsuperscript{3}\\
\vspace{8pt}
\;\;\textsuperscript{1}University of Bonn \quad\,
\textsuperscript{2}Lamarr Institute \quad\,
\textsuperscript{3}MPI for Informatics, SIC}

\begin{document}

\twocolumn[{%
\renewcommand\twocolumn[1][]{#1}%
\maketitle
\vspace{-20pt}
 \includegraphics[width=\linewidth]{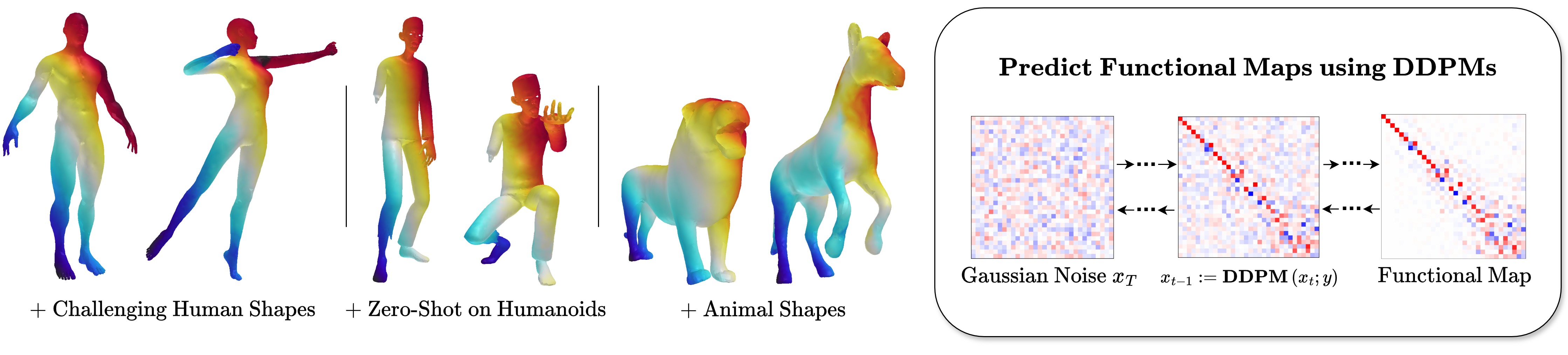}
\vspace{-2em}
\captionof{figure}{\textbf{We propose DenoisFM, a novel method for predicting shape correspondences in the form of functional maps using denoising diffusion models.} (Left:) Challenging examples our method can handle, with color-coded correspondences. (Right:) By sequentially denoising samples of random noise, the diffusion model can predict the correct functional map between a pair of shapes.\vspace{0.3cm}}
\label{fig:teaser}
}]

\begin{abstract}

Estimating correspondences between pairs of deformable shapes remains a challenging problem. 
Despite substantial progress, existing methods lack broad generalization capabilities and require category-specific training data.
To address these limitations, we propose a fundamentally new approach to shape correspondence based on denoising diffusion models. 
In our method, a diffusion model learns to directly predict the functional map, a low-dimensional representation of a point-wise map between shapes.
We use a large dataset of synthetic human meshes for training and employ two steps to reduce the number of functional maps that need to be learned.
First, the maps refer to a template rather than shape pairs.
Second, the functional map is defined in a basis of eigenvectors of the Laplacian, which is not unique due to sign ambiguity.
Therefore, we introduce an unsupervised approach to select a specific basis by correcting the signs of eigenvectors based on surface features.
Our model achieves competitive performance on standard human datasets, meshes with anisotropic connectivity, non-isometric humanoid shapes, as well as animals compared to existing descriptor-based and large-scale shape deformation methods. 
See our project page\footnote{\label{footnote:project_page}Project page: \url{https://alekseizhuravlev.github.io/denoising-functional-maps/}} for the source code\footnote{\label{footnote:source_code}Source code: \url{https://github.com/alekseizhuravlev/denoising-functional-maps/}} and the datasets. 

\end{abstract}

\section{Introduction}

A long-standing problem in computer graphics and vision is to find corresponding points in a pair of 3D shapes \cite{bronstein2010scale,ovsjanikov2012functional, eisenberger2020smooth, SeelbachBenkner2021, sun2023spatially, cao2023Unsupervised, marin2024nicp}. 
The point map can then be used for texture \cite{dinh2005texture}, segmentation \cite{rustamov2007laplace} and deformation transfer \cite{sumner2004deformation}, statistical shape analysis \cite{corman2015supervised}, and construction of parametric models \cite{SMPL:2015, Qian2020}. 
A popular approach to solving the correspondence problem involves the functional map framework \cite{ovsjanikov2012functional}, which represents the correspondence as a small matrix that maps between the eigenfunctions of the Laplace operator on each shape.
This representation is compact and allows to regularize constraints on the correspondences, such as bijectivity or area preservation.
The functional map framework has been extended in numerous works, such as matching shapes with partiality \cite{rodola2017partial,attaiki2021dpfm}, analyzing the properties of shape collections \cite{corman2015supervised} or improving accuracy through iterative refinement \cite{melzi2019zoomout,eisenberger2020smooth,ren2021discrete}.

Previous functional map approaches focused on using small datasets, rarely exceeding a hundred unique shapes.
This favored training models with lightweight architectures that could learn as much as possible with limited data, but these methods may face challenges with generalization to broad shape categories.
To develop a unified model that is applicable to various tasks, we draw attention to denoising diffusion models ~\cite{sohl2015deep,song2019generative,ho2020denoising}, which have impressive generalization ability when trained on large amounts of data.
In our work, we apply this approach to correspondence estimation using functional maps.

Based on the idea of sequentially denoising samples of random noise into samples from the data distribution, diffusion models have achieved unprecedented performance in high-fidelity, diverse, and controllable image synthesis \cite{po2024state}. 
The scalable U-Net architecture  \cite{ronneberger2015u} allows them to be trained on large datasets with hundreds of thousands of high-resolution images.
The application of diffusion models to shape matching has been limited so far, mainly focusing on feature extraction from pre-trained foundational models \cite{Dutt_2024_CVPR,abdelreheem2023zero}.
Interestingly, functional maps appear well-suited to bridge diffusion models and shape correspondence.
They are small-sized single-channel matrices, making standard architectures developed for images readily applicable to them.
Large amounts of shape data for training are then available through parametric models such as SMPL~\cite{SMPL:2015}, an approach that has already been used for matching tasks \cite{groueix20183d,trappolini2021shape} and is also suitable for diffusion models.

In this work, we present Denoising Functional Maps (DenoisFM), an approach to directly predict the functional map between shapes using a denoising diffusion model;  see Fig.~\ref{fig:teaser} for the overview. 
For training, we create a large dataset of functional maps between $2.3 \cdot 10^5$ human shapes from the SURREAL dataset \cite{varol2017learning} and a template shape.
Using template-wise maps instead of pair-wise ones allows to reduce the complexity of training and inference. 
The conditioning information is inferred directly from the geometric shape structure.
Additionally, we propose to exploit the probabilistic nature of diffusion models by predicting several initial maps instead of a single one, and selecting the best map based on the Dirichlet energy \cite{pinkall1993computing}. 

An important property of functional maps is that they are defined in the basis of the first $n$ eigenvectors of the Laplacian operator for each shape.
The basis is not unique due to the sign ambiguity of the eigenvectors, and there are potentially $2^n$ possible basis combinations for each shape.
To address this problem, we propose an unsupervised approach which selects the signs of the basis vectors using per-vertex features extracted in an unsupervised way (\cref{sec:sub:sign_ambiguity}).
This reduces the number of possible functional maps between shapes, further lowering the training complexity.

\vspace{-10pt}
\paragraph{Contributions. }
Overall, the technical contributions of this paper can be summarized as follows:
\begin{itemize} 
    \item DenoisFM, a fundamentally new approach to shape correspondence estimation based on predicting functional maps using denoising diffusion models; 
    \item An unsupervised method to address the sign ambiguity of eigenvectors of the Laplacian, which reduces the complexity of the prediction problem. 
\end{itemize}

Our model performs on par with existing shape matching methods on human benchmark datasets, achieves generalization to non-isometric humanoid shapes without learning on them, and is also applicable to animal shapes. 
See our project page\textsuperscript{\ref{footnote:project_page}} 
for the source code\textsuperscript{\ref{footnote:source_code}} and the datasets of functional maps used to train our models.

\section{Related Work}

\paragraph{Shape Correspondence Estimation.}
Learning-based shape correspondence methods can be roughly divided into two categories: first, approaches that predict point-wise descriptors suitable for computing functional maps \cite{ovsjanikov2012functional,ovsjanikov2016computing}, a low-rank representation of the point-wise correspondence in Laplace--Beltrami eigenbasis; second, deformation-based methods that directly align input shapes and typically require much larger training datasets than descriptor-based methods.

In the standard functional map pipeline, point-wise descriptor functions that encode the surface geometry are first computed on each shape \cite{aubry2011wave,bronstein2010scale,tombari2010unique}, and then the functional map is estimated using the descriptors in a least-squares optimization.
With the advent of deep learning, network architectures were introduced to learn descriptor functions directly from the data \cite{litany2017deep,sharp2022diffusionnet}.
While the initial methods were supervised \cite{litany2017deep,donati2020deep}, recent works estimate the descriptors in an unsupervised manner \cite{roufosse2019unsupervised,cao2023Unsupervised}.
Most shape matching datasets consist of fewer than a hundred unique meshes and focus on a specific category of shapes \cite{anguelov2005scape,bogo2014faust,melzi2019shrec,magnet2022smooth}.
Therefore, the most effective descriptor-based models use lightweight architectures to infer as much geometric information as possible from limited training data.

Another group of works use large-scale datasets of synthetic data for training.
3D-CODED \cite{groueix20183d} is an encoder-decoder architecture that predicts the correspondence by learning a surface parameterization and attempts to estimate the deformation from the template shape to the test one.
Transmatch \cite{trappolini2021shape} is a transformer-based model that uses an attention mechanism to learn the underlying structure of the geometry. 
Other works utilizing large amounts of data \cite{cao2023self,zeng2021corrnet3d,lang2021dpc} mainly focus on point cloud registration, and a group of works learn the shape deformation using data-efficient architectures \cite{eisenberger2021neuromorph,sundararaman2022implicit}.

We combine the advantages of both categories of models by training a large-scale method based on functional maps that leverages the power of denoising diffusion models.

\paragraph{Denoising Diffusion Models.}
Diffusion models have become popular in many areas of visual computing; see the recent survey for details \cite{po2024state}. 
Several diffusion-based generative models have been proposed \cite{sohl2015deep,song2019generative}, and the simplified learning process of denoising diffusion probabilistic models (DDPMs) enabled the image synthesis at unprecedented quality \cite{ho2020denoising}. 
Various conditioning mechanisms such as class labels, textual descriptions, or segmentation maps have been used to guide the diffusion process \cite{rombach2022high}. 
While diffusion models quickly found applications in areas beyond imaging such as video \cite{ho2022video,ho2022imagen} and 3D content generation \cite{poole2022dreamfusion,liu2023meshdiffusion}, their application 
to shape correspondence estimation has been limited so far to extracting descriptors from images of meshes \cite{Dutt_2024_CVPR} or segmentation maps \cite{abdelreheem2023zero} generated by foundational models. 
Our work is the first to use DDPMs to directly predict the functional maps 
between pairs of shapes to obtain the correspondences.

\paragraph{Sign Ambiguity of Laplacian Eigenvectors.}
Since for any eigenvector $\phi_i$, its negative version $-\phi_i$ is also an eigenvector, the Laplace-Beltrami eigenbasis has an inherent sign ambiguity.
Additionally, the eigenvalues can have higher multiplicity and make the intrinsic ordering non-unique.
This symmetry needs to be taken into account when processing the eigenvectors~\cite{rustamov2007laplace,ovsjanikov2008global,bro2008resolving}.

In graph learning, Laplacian eigenvectors are used to encode information about the structure of a graph \cite{belkin2003laplacian,levy2006laplace}, and several methods were proposed to address the sign ambiguity.
Laplacian Canonization~\cite{ma2024laplacian} proposes to find the coordinate axis on which the projection of the eigenvector has the largest angle and fixes the projection to be positive.
SAN~\cite{kreuzer2021rethinking} obtains sign-invariant operators using absolute subtraction and product between a pair of nodes, this only considers the relative sign between queries. 
SignNet \cite{lim2022sign} learns a sign-invariant embedding and Dwivedi et al.~\cite{dwivedi2023benchmarking} learns on all possible sign combinations. %

The methods described above are designed for graphs with a small number of vertices.
For point clouds, Linearly Invariant Embedding \cite{marin2020correspondence} considered the problem of learning the basis functions directly from the input data.
However, the majority of works on 3D shapes employ the sign-equivariant approach, where a sign-dependent functional map is computed from descriptor functions in a precomputed Laplace-Beltrami eigenbasis. 
In our work, we propose to address the sign ambiguity problem using a learning-based method that assigns a specific combination of signs to the eigenvectors of any human shape, which is learned directly from the data in an unsupervised way.

\section{Background}
\label{sec:background}

\paragraph{Notation.}
Given a shape $S$ represented as a triangular mesh with \( v \) vertices, we compute its Laplace matrix $L$ \cite{meyer2003discrete} and the first $n$ eigenvectors of $L$ corresponding to the eigenvalues $\lambda_1 \leqslant \ldots \leqslant \lambda_n$.
$A$ is the diagonal vertex-area matrix associated with $L$, computed by assigning one-third of the total area of all faces incident to the corresponding vertex.

We denote matrices of \( n \) functions on the vertices of \( S \) by uppercase letters: \( \Phi \in \mathbb{R}^{v \times n} \) for the matrix of eigenvectors, \( \Sigma \in \mathbb{R}^{v \times n} \) for the matrix of learned features.
The individual eigenvectors and feature vectors are denoted by lowercase letters: \( \phi_i \in \mathbb{R}^v \) for the \( i \)-th eigenvector, \( \varsigma_i \in \mathbb{R}^v \) for the \( i \)-th feature vector.
Finally, \( \sigma \in \{-1, 1\}^n \) represents a vector consisting of $n$ signs.

\subsection{Functional Maps}
\label{sec:sub:fmaps}

Our method is built upon the framework of functional
maps \cite{ovsjanikov2012functional, ovsjanikov2016computing}.
Consider the point-wise map $\Pi_{ 21 } \in\{0,1\}^{v_{ 2 } \times v_{ 1 }}$ between two meshes $S_1$ and $S_2$.
The point-wise map is a permutation matrix, and due to its large size, it cannot be processed efficiently.
The functional map $C_{ 1 2 } \in \mathbb{R} ^{n \times n}$ is the spectral representation of $\Pi_{ 21 }$, which is computed as 

\begin{equation}\label{eq:fmap}
    C_{ 1 2 }=\Phi_2^{\dagger} \Pi_{ 21 } \Phi_1,    
\end{equation}
\noindent where ``$^{\dagger}$'' denotes the pseudo-inverse.
The functional map efficiently encodes the point-wise map in a low-dimensional representation, since $n \ll v_1, v_2$.

In a standard pipeline, learned or handcrafted $p$-dimensional descriptor functions are computed on each shape, $F_1 \in \mathbb{R} ^{v_1 \times p}, F_2 \in \mathbb{R} ^{v_2 \times p}$.
The functional map is then obtained by solving the linear system \cite{ovsjanikov2016computing}:

\begin{equation}
    C_{ 12 }=\underset{ C }{\arg \min }\left\| C \Phi_1^{\dagger} F_1 - \Phi_2^{\dagger} F_2\right\|_2^2 ,
\end{equation}

\noindent which is usually further regularized \cite{donati2020deep,nogneng2017informative}.
The point-wise map can be recovered from the functional map and the eigenvectors using nearest neighbor search \cite{ovsjanikov2012functional,pai2021fast}.

\subsection{Sign Ambiguity}
\label{sec:sub:sign_ambiguity}

The functional map is defined in a basis spanned by the eigenvectors of the Laplacian on each shape, as can be seen from Eq.~\eqref{eq:fmap}.
However, this basis is not unique since for each eigenvector $\phi_i$, $- \phi_i$ is also a valid eigenvector corresponding to the same eigenvalue $\lambda_i$. 
Due to sign ambiguity, there are $2^{n}$ possible basis combinations on each shape.

The eigenvectors returned by a numerical eigensolver will have a random sign combination, and mathematically it is impossible to select some specific sign of eigenvectors.
However, several works demonstrate that the choice of the sign can be determined by the underlying data using the projection operator \cite{bro2008resolving,ma2024laplacian,lim2022sign}.
An intuitive example arises in the principal component analysis: Each eigenvector should point in the same direction as the points it is representing.
This can be achieved by projecting the singular vector onto the data vectors and changing its sign to make the projection positive.
This approach allows to select a canonical direction for the first few eigenvectors.

Sign ambiguity is a special case of basis ambiguity when all eigenvalues are distinct.
The more general basis ambiguity occurs if an eigenvalue $\lambda_i$ has multiplicity degree $d_i>1$.
Then any $d_i$ vectors forming an orthogonal basis in the subspace spanned by its eigenvectors $\left(\phi_{i+1}, \phi_{i+2} \ldots \phi_{i+d_{i}} \right)$ are also valid eigenvectors.
Although the basis ambiguity is difficult to deal with, in practice high multiplicities do not occur often in low-order eigenvectors \cite{kato2013perturbation,ovsjanikov2016computing}.

\subsection{Denoising Diffusion Models}
Given a dataset of examples drawn independently from a data distribution $q ( x )$, diffusion models aim to learn the underlying data distribution by sequentially denoising random noise samples \cite{sohl2015deep,ho2020denoising,po2024state}.

The forward diffusion process takes a real data point $x _{ 0 }$ and incrementally adds to it a small amount of Gaussian noise, producing a sequence of $T$ noisy samples $x _1, \ldots, x _T$. 
The step sizes are controlled by a variance schedule $\left\{\beta_t \in(0,1)\right\}_{t=1}^T$. 
With the larger steps $t$, the noisy sample $x _t$ further deviates from the original sample. 
At $T\rightarrow\infty$, it is close to the Gaussian distribution, $x_T \sim N \left(x_T ; 0, I\right)$.
The reverse process recreates the original sample by iteratively removing noise from an input $x_T$ drawn from the Gaussian distribution. 
Assuming that the reverse conditional probability is conditioned on $x _{ 0 }$, $q \left( x _{t-1} \mid x _t, x _{ 0 }\right)= N \left( x _{t-1} ; \tilde{ \mu }\left( x _t, x _0\right), \tilde{\beta}_t I \right),$
we can obtain 
$\tilde{ \mu }_t =\frac{1}{\sqrt{\alpha_t}}\left( x _t-\frac{1-\alpha_t}{\sqrt{1-\bar{\alpha}_t}} \epsilon _t\right)$,
where $\alpha_t=1-\beta_t$ and $\bar{\alpha}_t=\prod_{i=1}^T \alpha_i$.

Instead of directly predicting the denoised sample $\tilde{ \mu }_t$ with a neural network, we can predict the noise content $\epsilon_\theta\left( x _t, t\right)$ from the input $x _t$ at time step $t$ \cite{ho2020denoising}.
Such a model is trained with a denoising score-matching objective:
$\mathcal{L}_t = E _{t \sim[1, T], x _0, \epsilon_t}\left[\left\| \epsilon _t- \epsilon _\theta\left( x _t, t\right)\right\|^2\right].$

\section{Denoising Functional Maps} 
\label{sec:method}

\begin{figure*}[ht]
\centering
\begin{subfigure}[b]{\linewidth}
\centering
   \includegraphics[width=0.95\linewidth]{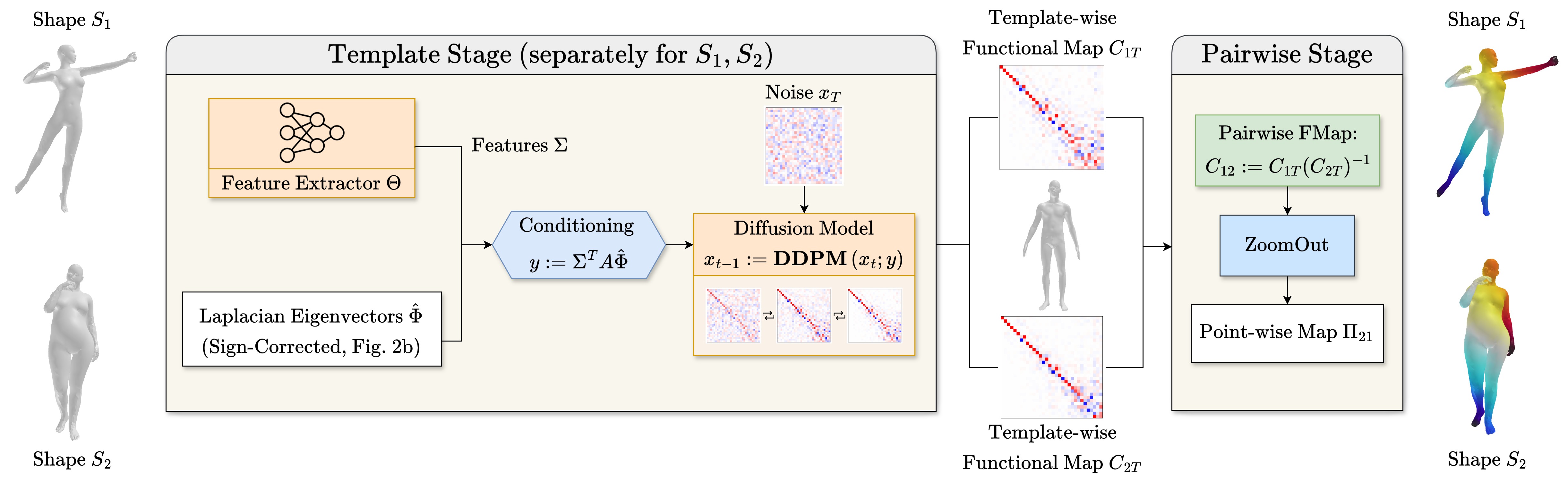}
   \caption{\textbf{Functional map prediction using a diffusion model}. For each unique shape $S$ in the dataset, we obtain feature vectors $\Sigma$ using a feature extractor $\Theta$, as well as a sign-corrected eigenbasis $\hat{\Phi}$ (Fig.~\ref{fig:Ng2}, Sec.~\ref{ssec:sign_correction_part}). Both are used for conditional prediction of the template-wise functional map $C_{1T}$ using a diffusion model, performed separately for each shape. The pairwise functional maps $C_{12}$ are obtained through the map composition property, upsampled using ZoomOut \cite{melzi2019zoomout}, and converted to pointwise maps $\Pi_{21}$. 
   }
   \hspace{10mm}
   \label{fig:Ng1} 
\end{subfigure}
\centering
\begin{subfigure}[b]{\linewidth}
\centering
   \includegraphics[width=0.8\linewidth]{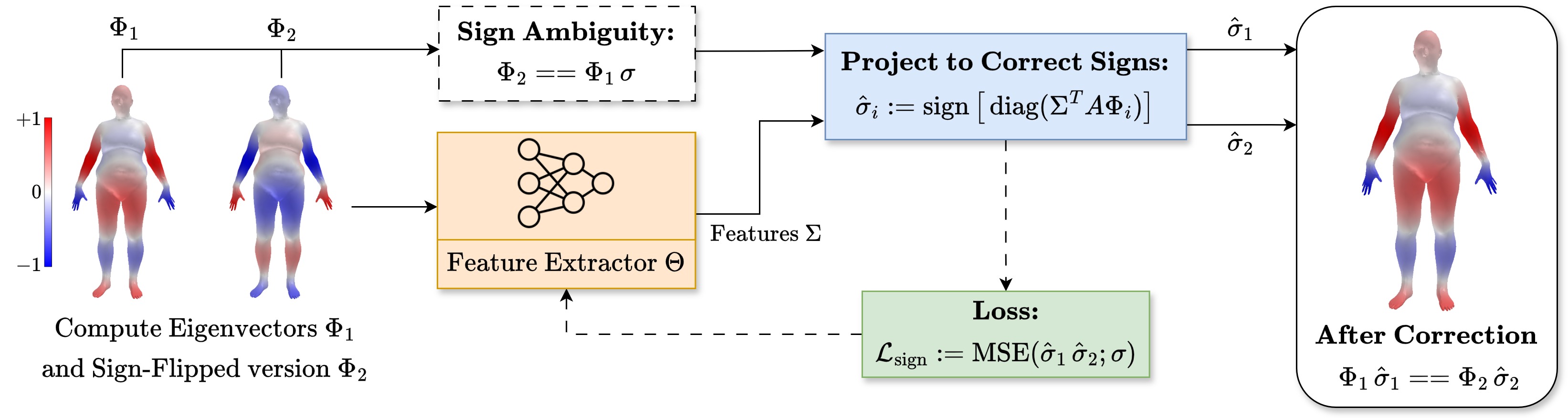}
   \caption{\textbf{Learned sign correction}. The eigenvectors $\Phi$ returned by a numerical eigensolver have random signs. To select a specific sign, we project the eigenvectors onto the feature vectors $\Sigma$ obtained with a feature extractor $\Theta$, and make the projection positive. The sign correction network is trained in an unsupervised manner: learning to correct eigenvectors that have random signs. The learned features $\Sigma$ are also used for the conditional prediction of the template-wise functional map using a diffusion model (Fig.~\ref{fig:Ng1}).}
   \label{fig:Ng2}
\end{subfigure}
\caption{\textbf{Overview of DenoisFM.} (a) A pair of shapes is matched using a DDPM model. (b) The signs of eigenvectors are corrected through learned features.}
\label{fig:method_1} 
\end{figure*}

This section introduces Denoising Functional Maps (\mbox{DenoisFM}), our approach for shape correspondence estimation; see Fig.~\ref{fig:method_1} for the method overview. 
Since functional maps are square low-dimensional matrices (see Sec.~\ref{sec:sub:fmaps}), 
we use a standard neural architecture based on a 2D U-Net~\cite{unet} as a backbone. 
We aim to train our model on a large collection of shapes with diverse geometry. 
Since functional maps are defined for shape pairs, for a dataset with $m$ shapes the model would have to learn $m^2$ functional maps.
Instead of this naive approach, we consider the correspondences to a single template shape, following previous works \cite{groueix20183d,trappolini2021shape}. 
As a result, the training data includes only $m$ functional maps, one for each shape in the dataset. 

A functional map is defined in a specific basis of Laplacian eigenvectors, and knowing the exact basis combination is necessary to convert it to a point-wise correspondence map. 
Therefore, we compute the basis first and then condition the diffusion model on it; the
model then returns the corresponding functional map.
To reduce the complexity of the problem caused by the sign ambiguity of basis eigenvectors, we utilize a sign correction network to select a specific combination of signs for the basis.

DenoisFM consists of two components: the DDPM for functional map prediction (Sec.~\ref{ssec:DDPM_part}) with post-processing that filters out low-quality solutions using Dirichlet energy of the point-wise map (Sec.~\ref{sec:sub:dirichlet_selection}), and the sign corrector for the Laplacian eigenvectors (Sec.~\ref{ssec:sign_correction_part}).

\subsection{Denoising Diffusion Component} 
\label{ssec:DDPM_part} 

We next describe the training and inference of the diffusion component of DenoisFM. 
For training, we use a large collection of shapes and a template.
For each shape $S$ in the dataset, we compute an eigenbasis $\Phi$, a template-wise functional map $C_{1T}$, and conditioning $y = y\left(S, \Phi\right)$ which incorporates geometric details of the input shape and information about the basis. 
The exact form of conditioning will be specified in Sec. \ref{sec:sub:conditioning}.
We use the training objective of denoising diffusion probabilistic models (DDPMs) \cite{ho2020denoising}:
$\mathcal{L}_{\text{DDPM}}= E _{x, \epsilon \sim N (0,1), t}\left[\left\|\epsilon-\epsilon_\theta\left(x_t, t, y\right)\right\|_2^2\right],$
where $\epsilon_\theta$ is the amount of noise predicted by a 2D UNet \cite{ronneberger2015u}, $x_t$ is a noisy version of the input $x_0 \coloneq C_{1T}$, and 
$t$ is the time step uniformly sampled from $\{1, \ldots, T\}$. 

The inference pipeline consists of two stages, see Fig.~\ref{fig:Ng1}.
In the template stage, for each individual shape $S$ in the evaluation dataset, we obtain an eigenbasis $\Phi$, calculate the conditioning $y$, and predict the template-wise functional map $C_{1T}$ using the diffusion model.
Note that the inference of DDPMs is time-consuming due to the large number of denoising steps and can take several hours for high-dimensional functional maps.
Because the predicted maps refer to the template, we need to run the denoising process only once for each shape. 

In the pairwise stage, for each test pair, we convert the respective template-wise functional maps into pairwise ones using the composition property \cite{ovsjanikov2012functional}: $C_{12} = C_{1T} (C_{2T})^{-1}$.
We then apply the Zoomout refinement \cite{melzi2019zoomout} to increase the resolution of the functional maps. 
This spectral upsampling step has been shown to improve their accuracy, and our framework benefits from it without additional modifications.
Finally, the functional map is converted into a point-wise map $\Pi_{21}$ using nearest neighbor search.

\subsubsection{Post-Processing of the Samples}
\label{sec:sub:dirichlet_selection}

Diffusion models are inherently probabilistic and their outputs may differ given the same conditioning but different initial noise \cite{ho2020denoising}.
To address this, we repeat the denoising process multiple times and filter out the low-quality results. 
The selection is based on the Dirichlet energy of the point-wise map \cite{magnet2022smooth, ezuz2019reversible}: an accurate map should be smooth, with neighboring vertices on one shape matching neighboring vertices on the other.
Therefore, we repeat both the template step and the pairwise step to produce multiple point-wise maps.
We can then either report the map with the lowest Dirichlet energy or combine the results of several smoothest maps.
To do this, for each query point on the source mesh, we gather potential matching points on the target mesh and select the one with the smallest total distance to the other candidates (see Supp.~\ref{ssec:map_selection_supp}).
The map selection step has a major impact on the accuracy of our method.

\subsection{Learned Sign Correction} 
\label{ssec:sign_correction_part} 

The diffusion model needs to be conditioned on the basis eigenvectors and predict the functional map for that particular basis. 
Theoretically, the model could be trained using random basis combinations, as was done in Dwivedi \textit{et al.}~\cite{dwivedi2023benchmarking} for graph networks. 
However, this would significantly increase the amount of training data. 

Hence, we follow a more effective approach:
we select a specific basis by fixing the sign of the Laplacian eigenvectors on each shape. 
Previous works infer the sign from the underlying data using the projection operator (see Sec.~\ref{sec:sub:sign_ambiguity}). 
Therefore, suppose that $\Sigma := \Theta\left(S\right) \in \mathbb{R}^{v\times n}$ is a set of per-vertex feature vectors inferred from the geometry of shape $S$ by a trainable feature extractor $\Theta$, one for each eigenvector.
The sign of the eigenvectors can be selected by projecting the features $\Sigma$ onto the eigenvectors $\Phi$ and making the corresponding projections positive:

\begin{equation}
\label{eq:sign_correction}
\begin{aligned}
    \hat{\sigma} &\coloneq \operatorname{sign}\big[\operatorname{diag}(\Sigma^T A \Phi)\big],\\
    \hat{\Phi} &\coloneq \Phi\,\hat{\sigma}, 
\end{aligned}  
\end{equation}
where ``$\operatorname{diag}(\cdot)$'' and ``$\operatorname{sign}(\cdot)$'' denote matrix diagonal and sign extraction operators, $A$ is the diagonal vertex-area matrix. 
To illustrate, suppose we have a human mesh and a single feature vector $\varsigma_i$, which is a binary vector with ones for points on the left hand, and zeros otherwise. 
After performing the procedure described in Eq.~\eqref{eq:sign_correction}, the corrected eigenvector $\hat{\phi}_i$ will be mostly positive on the left hand.

\subsubsection{Training the Sign Corrector}
\label{ssec:training_sign_corr}

The sign correction network is trained in an unsupervised manner by learning to correct eigenvectors that have random signs.
If we perform the eigendecomposition with a numerical solver twice on the same mesh and get two sets of eigenvectors $\Phi_1$ and $\Phi_2$, due to the sign ambiguity, the difference between them can be written as $\Phi_2 = \Phi_1 \, \sigma$,
where $\sigma \in \{-1, 1\}^n$ is the ground-truth sign difference. 
Now we perform the sign correction using Eq.~\eqref{eq:sign_correction}, after which we expect the corrected eigenvectors to be equal to each other: 
\begin{equation}
\begin{aligned}
    \hat{\Phi}_1 &= \hat{\Phi}_2, \\
    \Phi_1 \, \hat{\sigma}_1 &= \Phi_2 \, \hat{\sigma}_2,
\end{aligned}
\end{equation}
where $\hat{\sigma}_1, \hat{\sigma}_2 \in \{-1, 1\}^n$ are the predicted combinations of correction signs.
After substitution and using $\hat{\sigma}_1 \, \hat{\sigma}_1 = I$:
\begin{equation}
\begin{aligned}
    \Phi_1 \, \hat{\sigma}_1 &= \Phi_1 \, \sigma \, \hat{\sigma}_2, \\
    \hat{\sigma}_1 &= \sigma \, \hat{\sigma}_2, \\
    \hat{\sigma}_1 \, \hat{\sigma}_2 &= \sigma.
\end{aligned}
\end{equation}

\noindent 
As a result, %
the difference between the predicted correction signs should be the same as the ground-truth sign difference, and we can use this as our unsupervised training objective with a mean square error function:

\begin{equation}
\label{eq:objective_sign_corr}
    \mathcal{L}_{\operatorname{sign}} = \operatorname{MSE}\left(\hat{\sigma}_1 \hat{\sigma}_2 ; \sigma\right)
\end{equation}

\noindent Note that the $\operatorname{sign}$ operator in Eq.~\eqref{eq:sign_correction} is not differentiable. 
Therefore, during training, we set $\hat{\sigma}_1$ and $\hat{\sigma}_2$ to be the diagonal elements of the projections, rather than their signs, assuming that the feature vectors and eigenvectors have unit norm. 
The training objective of the feature extractor (Eq.~\ref{eq:objective_sign_corr}) then forces the elements of $\hat{\sigma}_1$ and $\hat{\sigma}_2$ to be as close to $-1$ or $1$ as possible. 
The absolute value can be interpreted as confidence in the predicted sign: values close to $\pm1$ mean high confidence.

In practice, we found that a network trained this way quickly converges on a dataset of human or animal meshes. 
Our method returns the same corrected eigenbasis 95--99\% of the time for the first hundred Laplacian eigenvectors. 
The accuracy of the sign correction and its impact on the distribution of functional maps are discussed in Supp.~\ref{sec:supp:sign_corr}.

\subsection{Choice of Conditioning}
\label{sec:sub:conditioning}

Conditioning of the diffusion model should include two parts: the basis information and the geometric properties of the shape. 
The former can be provided simply via the sign-corrected eigenvectors $\hat{\Phi}$. 
They additionally represent the spectral embedding of the shape and convey the full intrinsic information.
As for the latter, during the training of the sign-correction network, the useful properties will be encoded in the learned feature vectors $\Sigma$. 
Therefore, we propose to project the feature vectors onto the sign-corrected eigenvectors: 
\begin{equation}
    \label{eq:conditioning}
    \textbf{Conditioning } y = \Sigma^T A \hat{\Phi}. 
\end{equation}
This form of conditioning has two important properties. 
First, it has the same dimensions as the functional map.
Therefore, we can condition the diffusion model using a simple concatenation, where the conditioning matrix is concatenated with the intermediate denoising targets and passed through the score estimator as input \cite{ho2020denoising,weng2021diffusion}. 
Second, it is independent of the number of vertices, making it scalable to large meshes.

\section{Training and Experimental Results}
\label{sec:results}
This section describes the steps for generating the training data for our model (Sec.~\ref{sec:sub:augmentation}) as well as the experimental results on several challenging datasets (Secs.~\ref{ssec:nearisometric}--\ref{sec:sub:animals}). 
The evaluation of the sign correction network, additional comparison, implementation details, and the ablation studies can be found in Supp.~\ref{sec:supp:sign_corr} to \ref{sec:supp:add_eval}.
Since our model is designed to be trained on large and diverse shape collections, it can be applied to multiple (reminiscent) shape classes without re-training. 
This reduces the need for small category-specific training sets. 
We focus on datasets with human-like shapes as their combined volume allows versatile comparison of our approach with the previous works.

\subsection{Training Data and Implementation}

\label{sec:sub:augmentation}

Our model consists of two independent components: a denoising diffusion model that predicts the functional map and a sign correction network based on a per-vertex feature extractor, trained separately on their respective datasets.

\subsubsection{Denoising Diffusion}\label{ssec:DenoisingDiffusion}

Diffusion models require large amounts of training data with hundreds of thousands of examples. 
Therefore, we follow 3D-CODED \cite{groueix20183d} and use the SURREAL \cite{varol2017learning} dataset with 230,000 human meshes generated with SMPL \cite{SMPL:2015}. 
All its meshes are smooth and have the same vertex structure. 
To make the connectivity more diverse, we remesh each mesh \cite{pymeshlab,hoppe1993mesh} and randomly simplify the number of faces using Quadric Edge Collapse \cite{garland1997surface}. 
In $65\%$ of the cases, we simplify the entire mesh resulting in uniform connectivity. 
In the remaining $35\%$ of the cases, we simplify only a small subset of the mesh to add non-even connectivity to the training data. 
After remeshing, we correct the signs of the eigenvectors using a pre-trained sign correction network (Sec. \ref{ssec:sign_correction_part}) and compute the functional map to the template shape. 
We only store the functional maps and conditioning matrices to save memory requirements. 
The data generation takes 10 hours using parallel processing.

We use the standard DDPM \cite{ho2020denoising} implemented in Diffusers \cite{von-platen-etal-2022-diffusers} and distributed training with Accelerate \cite{accelerate}. 
We train models that predict 32, 64, and 96-dimensional functional maps for 100 epochs on eight A40 GPUs, which takes 1, 8, and 30 hours respectively. 
The inference at the template stage takes 1, 7, and 27 minutes for each shape in the dataset. 
The pairwise stage takes 30 seconds per pair.

\subsubsection{Sign Correction Network}
\label{sec:sub:implementation_sign_corr}

The per-vertex feature extractor of the sign correction network is based on DiffusionNet \cite{sharp2022diffusionnet} (not to be confused with denoising diffusion) with six blocks and 128 WKS descriptors \cite{aubry2011wave} as input.
This architecture was designed for small shape matching datasets, so we used 1000 randomly selected meshes from SURREAL for training. 
Each of them was randomly remeshed and simplified without adding anisotropy. 
The sign correction network is trained for $5 \cdot 10^4$ iterations, which takes one hour on a single A40 GPU; the inference takes a few seconds per shape.

\subsection{Near-Isometric Shape Matching}
\label{ssec:nearisometric}

\paragraph{Datasets.} 
We evaluate our method on three widely used datasets: FAUST \cite{bogo2014faust}, SCAPE \cite{anguelov2005scape} and SHREC'19 \cite{melzi2019shrec}. 
Following previous works, we use their more challenging remeshed versions \cite{donati2020deep,ren2018continuous}: 
\begin{itemize}
    \item The \textbf{FAUST} dataset contains meshes of 10 human classes in 10 poses, 100 in total. The train/test split is 80/20.
    \item The \textbf{SCAPE} dataset includes 71 meshes of the same person in different poses, the last 20 are used for testing.
    \item The \textbf{SHREC'19} dataset is composed of 44 shapes with diverse body types and poses, arguably the most challenging of the three. 
    This dataset does not have a train split and is used for testing only.
    Following \cite{cao2023Unsupervised}, we removed shape 40 since it is the only partial non-closed shape.
\end{itemize}

\paragraph{Baselines.}
We compare our approach with two categories of non-rigid shape correspondence methods. 
The first includes large-scale template-based models trained on SURREAL: 3D-CODED \cite{groueix20183d} and TransMatch \cite{trappolini2021shape}, i.e.,~our main competitors as our work belongs to the same group.
The second category includes descriptor-based models: ULRSSM \cite{cao2023Unsupervised}, DiffZO \cite{magnet2024memory}, ConsistentFMaps \cite{sun2023spatially}, GeomFMaps \cite{donati2020deep}, AttentiveFMaps \cite{li2022learning}, DUO-FMNet \cite{donati2022deep_orient}, SSL \cite{cao2023self}, SmS \cite{cao2024spectral}. 
Since our and other methods can be applied to any shape class, we do not include template fitting methods such as NICP \cite{marin2024nicp}, which typically focus on human scans and incorporate explicit human body priors.

To compare both categories of baselines fairly, we note that SCAPE is too small to train large models. 
On the other hand, FAUST was created with SMPL \cite{SMPL:2015} like SURREAL, and includes a sufficient diversity of body types and poses. 
Therefore, we follow \cite{cao2023Unsupervised,sun2023spatially,cao2023self} and train large models on SURREAL, descriptor-based models on FAUST, and evaluate them on three test datasets. 
A comparison with the baselines trained on FAUST+SCAPE can be found in Supp.~\ref{sec:results_baselines_faust_scape}.

\begin{table}[]
    \small
    \centering
    \begin{tabular}{lcccc}
    \hline \textbf{Model} & \textbf{Category} & \textbf{F\_r} & \textbf{S\_r} & \textbf{S'19\_r} \\
    \hline
    3D-CODED & \multirow{2}{*}{\shortstack{Large-\\Scale}} & 2.5 & 16.1 & 17.3 \\
    TransMatch &  & 1.7 & 15.3 & 21.0 \\
    \hline 
    DUO-FMNet & \multirow{8}{*}{\shortstack{Descriptor-\\Based}}& 2.5 & 4.2 & 6.4 \\
    GeomFMaps & & 1.9 & 2.4 & 7.9 \\
    AttentiveFMaps & & 1.9 & 2.6 & 5.8 \\
    ConsistentFMaps &  & 2.3 & 2.6 & 3.8 \\
    SSL& & 2.0& 3.1&4.0\\
    DiffZO & & 1.9 & 2.4 & 4.2 \\
    ULRSSM &  & 1.6 & 2.2 & 5.7 \\
    SmS& & \textbf{1.4}& 3.3&6.2\\
    \hline 
    \textbf{Ours} -- 64$\times$64 & & 1.8 & 2.3 & \textbf{3.5} \\
    \textbf{Ours} -- 96$\times$96 & & 1.7 & \textbf{2.1} & 3.9 \\
    
    \hline
    \end{tabular}
    \caption{Mean geodesic errors $\left(\times100\right)$ on the remeshed FAUST, SCAPE, and SHREC'19 datasets. The \textbf{best} results are highlighted. Large-scale methods are trained on SURREAL, and descriptor-based methods on FAUST. Here and in the following, the results of our model are averaged across 5 runs.}
    \label{tab:near_isometric}
\end{table}

\begin{figure}[t]
    \centering
    \includegraphics[width=\linewidth]{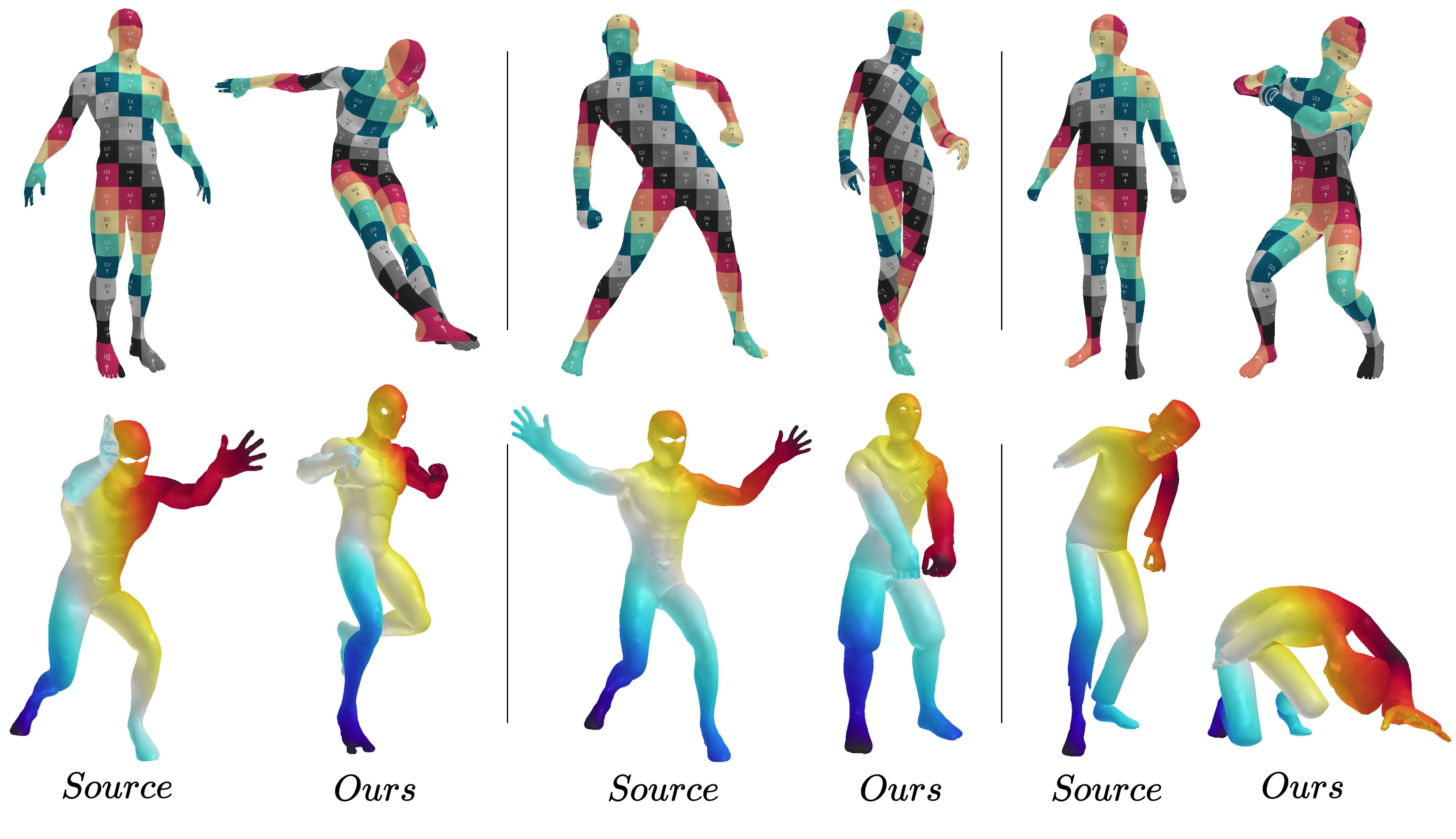}
    \caption{Qualitative results by our approach on SHREC'19 (top) and DT4D (bottom) datasets; 
    see also Supp.~\ref{sec:qualitative_supp}.
    } 
    \vspace{-6pt}
    \label{fig:qualitative}
\end{figure}

\paragraph{Results.}
We report the mean geodesic error of our model and the baselines in Table \ref{tab:near_isometric}. 
Compared to other large-scale methods trained on SURREAL, our work generalizes well to meshes from SCAPE and SHREC'19. 
A common problem in large-scale baselines is predicted point maps with incorrect symmetry.
This is caused by the fact that the geometry of the evaluation meshes has fewer fine-scale details such as fingers, which help to disambiguate the internal symmetry. 
Our work avoids overfitting to fine-scale details of SURREAL by using data augmentation (Sec.~\ref{sec:sub:augmentation}). 
As a result, we can utilize large amounts of data more effectively.

Compared to descriptor-based methods, our results are on par on all datasets (especially SHREC'19), demonstrating our excellent performance on diverse shape classes.
We can see that most descriptor-based methods can generalize to non-SMPL meshes, but none of them show competitive performance on all three datasets simultaneously.

We show the qualitative results of our model on several challenging meshes from SHREC'19 in Fig.~\ref{fig:qualitative}. 
Even though these shapes include high-frequency details such as muscles, in contrast to the smooth shapes typically generated by SMPL, our model is able to obtain correct correspondences.

\subsection{Matching with Anisotropic Meshing}
\label{ssec:anisotropic}

\paragraph{Datasets.}
To evaluate the robustness of our method to different connectivity, we use the anisotropically remeshed versions of FAUST and SCAPE \cite{donati2022deep_orient}, denoted F\_a and S\_a, respectively. 
In these datasets, the triangle scale is uneven, ranging from small and densely packed triangles on one side of the mesh to large and coarse triangulation on the other. 
As a result, models that are sensitive to mesh connectivity will likely fail to predict the correct maps. 

\begin{table}[]
    \centering
    \small
    \begin{tabular}[b]{lclc}
    \hline \textbf{Model} & \textbf{Category} & \textbf{F\_a} & \textbf{S\_a} \\
    \hline 
    3D-CODED & \multirow{2}{*}{\shortstack{Large\\Scale}} & 2.9 & 16.9 \\
    TransMatch &  & 2.7 & 16.2 \\
    \hline 
    DUO-FMNet & \multirow{5}{*}{\shortstack{Descriptor\\Based}} & 3.0 & 4.4 \\
    GeomFMaps & & 3.2 & 3.8 \\
    AttentiveFMaps &  & 2.4 & 2.8 \\
    ConsistentFMaps &  & 2.6 & 2.7 \\
    ULRSSM &  & \textbf{1.9} & 2.4 \\
    \hline 
    \textbf{Ours} - 64$\times$64 & & 2.1 & 2.3 \\
    \textbf{Ours} - 96$\times$96 & & 2.0 & \textbf{2.2} \\
    
    \hline
    \end{tabular}
    \caption{
    Results on the anisotropic FAUST and SCAPE.
    }
    \label{tab:anisotropic}
\end{table}

\paragraph{Results.}
Table \ref{tab:anisotropic} shows the matching results. Both large-scale and descriptor-based methods experience performance drops on anisotropically remeshed datasets. 
In contrast, our method is more robust to changes in mesh connectivity.
We attribute this to our design choices, such as diversifying the vertex structure of the training data and using conditioning that is independent of the number of vertices.

\subsection{Zero-Shot Shape Matching}
\label{ssec:zeroshot}

\paragraph{Datasets.}
Generalization to shape types not observed during training is a desirable property in many shape matching applications. 
To evaluate our model on unseen data, we use the non-isometric DT4D dataset \cite{magnet2022smooth} that consists of nine humanoid shape classes. 
While typically, 198 shapes are defined for training and 95 for testing, we instead make the evaluation more challenging by training on human datasets, SURREAL or FAUST. 
This can be viewed as a zero-shot setting, since most of the test shapes are different from the training data (see Supp.~\ref{sec:qualitative_supp} for an illustration).

\begin{table}[]
\small
    \centering
    \begin{tabular}{l|ccc|ccc}
    \hline
    \multirow{2}{*}{\textbf{Class}} & \multicolumn{3}{c|}{\textbf{Intra}} & \multicolumn{3}{c}{\textbf{Inter}} \\
    & \textbf{Ours} & \cite{cao2023Unsupervised} & \cite{sun2023spatially}& \textbf{Ours} & \cite{cao2023Unsupervised}  & \cite{sun2023spatially}\\
    \hline
manneq. & 1.0 & 1.0&1.6& \textbf{3.3} & 4.0& 4.5\\
zlorp & \textbf{1.1} & 1.3&2.1& \textbf{4.2} & 6.9 & \textbf{4.2}\\
crypto & \textbf{1.1} & 1.5&2.4& -- & -- & --\\
prisoner & 1.1 & 1.1&2.1& 29.6 & 20.5 & 26.7\\
ninja & \textbf{1.4} & 5.6&2.2& \textbf{4.3} & 16.2 & 7.1\\
ortiz & 9.3 & 4.4&\textbf{2.5}& -- & -- & -- \\
mousey & 10.2 & \textbf{2.7}&5.2& -- & -- & -- \\
drake & 10.6 & \textbf{1.0}&2.1& 10.4 & 8.2  & \textbf{4.5}\\
skeletonz. & 16.3 & 1.4&1.4& 49.4 & 36.5 & 48.0\\
    \hline
    \end{tabular}
    \caption{
    Results on shape classes from DT4D. 
    Classes \textit{mousey} and \textit{ortiz} are not present in inter-class split.
    All inter-class evaluation pairs have the format ``\textit{crypto} vs other class", so \textit{crypto} is excluded.
    }
    \vspace{-6pt}
    \label{tab:d4td}
\end{table}

\paragraph{Results.}
Our proposed experimental setup is challenging due to the differences between the training and test shapes. 
We use a lower-resolution model predicting a 32$\times$32 functional map, results of all our models are shown in Supp.~\ref{sec:supp:d4td_basis_number}. 
We compare our method with recent descriptor-based approaches: ConsistentFMaps \cite{sun2023spatially} and ULRSSM \cite{cao2023Unsupervised}.

The mean geodesic errors per class for inter- and intra-category splits are shown in Table \ref{tab:d4td}, with further comparisons in Supp.~\ref{sec:ulrssm_dt4d_supp}. 
Our method achieves accurate matching for most shape classes in both splits, demonstrating the generalization of our method to data unseen during training. 
We show the qualitative examples in Fig.~\ref{fig:qualitative}. 

\subsection{Application to Animal Shapes}
\label{sec:sub:animals}

\paragraph{Datasets.} 
To demonstrate the applicability of our method to shape classes other than humans, we test our approach on the SMAL dataset containing seven animal types.
We take the 32/17 split from \cite{donati2022deep_orient}, where all categories are present in train/test sets.
While the original training set is sufficient for descriptor-based baselines, it is too small for our model.
To address this, we use the fitted parameters of the SMAL parametric model \cite{zuffi20173d} and randomly vary the pose of each training animal, generating a total of 64,000 meshes.
We use the mean SMAL shape as a template.
Note that this generated dataset conveys little new information, since e.g., there are only three different types of lions.

\paragraph{Results.}
Table \ref{tab:smal} shows the results of our 64$\times$64 model.
Although the training data for our method lacks sufficient diversity, we achieve comparable performance to the baselines.
Therefore, given enough training data, we believe our method can be applied to any shape class.

\begin{table}[t]
    \begin{minipage}{.55\linewidth}
      \centering
      \small
        \begin{tabular}{l|ccc}
        \hline
        \textbf{Class} & \textbf{Ours} & \cite{cao2023Unsupervised} & \cite{sun2023spatially} \\
        \hline
        wolf & 3.5 & 3.5& 4.4 \\
        dog & 3.5 & 3.5& 4.5 \\
        horse & \textbf{3.7} & 3.8 & 4.4 \\
        cow & 3.8 & 3.8& 4.4 \\
        lion & 4.0 & \textbf{3.8}& 4.9 \\
        fox & 4.7 & \textbf{3.6}& 4.9 \\
        hippo & \textbf{7.6} & 9.9 & 7.9 \\
        \hline
        mean & 4.3 & 4.3& 4.8 \\
        \hline
        \end{tabular}
    \end{minipage}%
    \begin{minipage}{.4\linewidth}
      \centering
        \includegraphics[width=\linewidth]{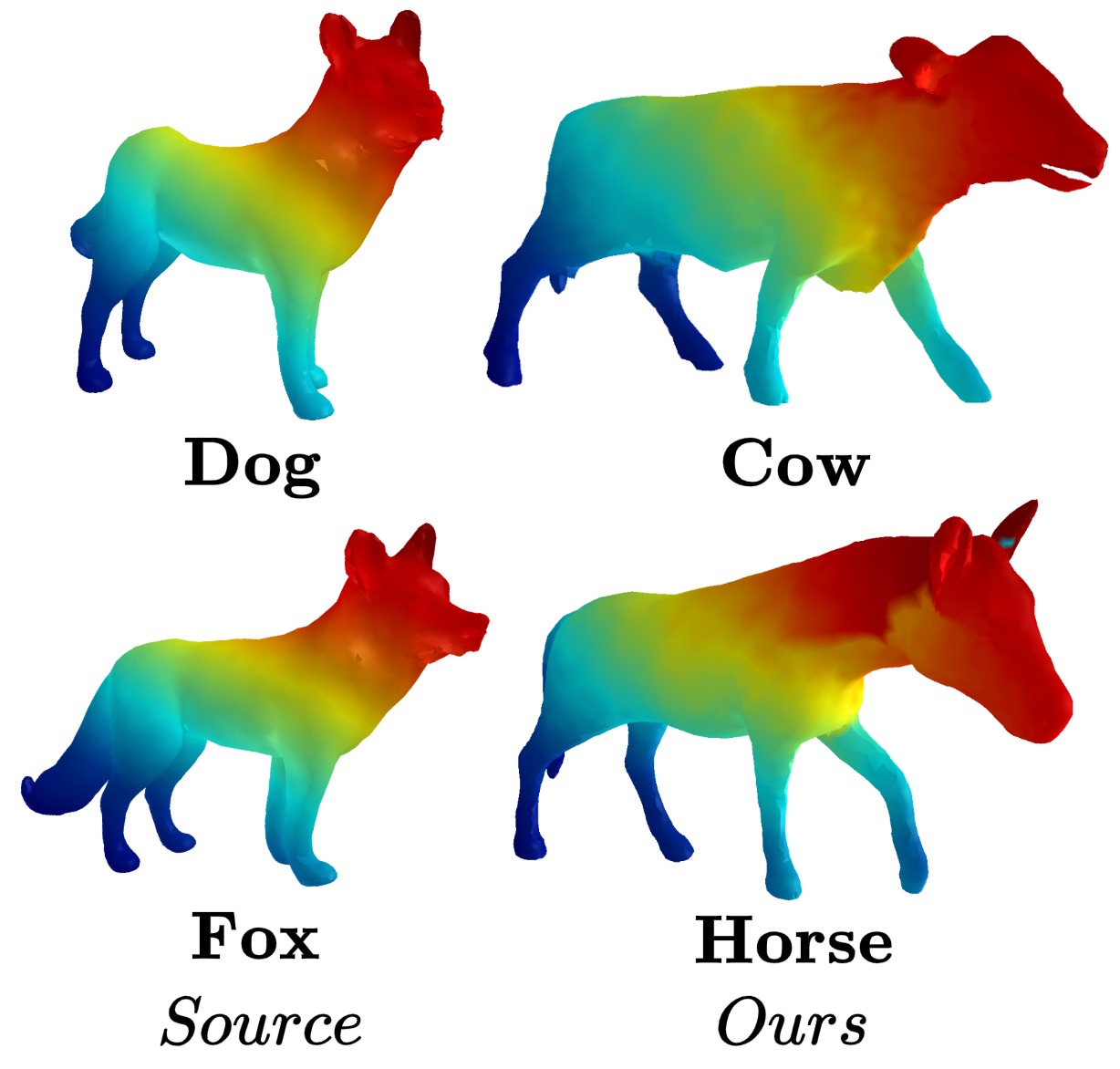}
    \end{minipage} 
    \caption{
    Results on the SMAL dataset \cite{donati2022deep_orient}, all categories are present in train/test sets. 
    Format: \textit{category} vs. everything.
    }
    \vspace{-6pt}
    \label{tab:smal}
\end{table}

\section{Conclusion}

We presented DenoisFM, an approach of a new class for shape correspondence estimation.
Our model is trained on a large dataset of synthetic human meshes and predicts the functional map between a pair of shapes using a denoising diffusion model.
DenoisFM achieves correspondence accuracy on par with state-of-the-art methods across multiple datasets.
The key advantage of our work is its generalization ability to human and humanoid shapes that differ noticeably from the training data, as well as competitive performance on animal shapes.
Therefore, DenoisFM bridges the gap between previous large-scale and descriptor-based approaches by efficiently leveraging large amounts of data, opening up a new direction for using diffusion models in shape correspondence estimation.

\section*{Acknowledgement} 
ZL acknowledges the support of the DFG Sachbeihilfe grant LA 5191/2-1.

{
    \small
    \bibliographystyle{ieeenat_fullname}
    \bibliography{main}
}

\clearpage

\twocolumn[{%
\renewcommand\twocolumn[1][]{#1}%
\maketitlesupplementary
 \includegraphics[width=\linewidth]{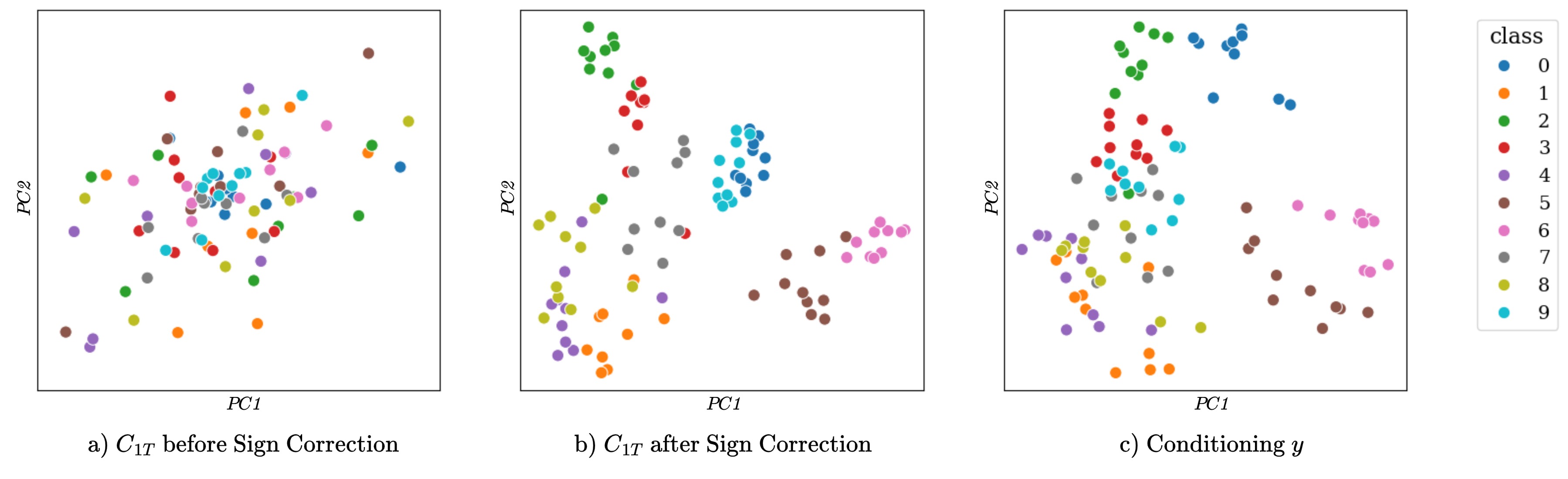}
\captionof{figure}{For the FAUST \cite{bogo2014faust} dataset, we plot the first two PCA components for: a) the functional maps between each shape and the template before sign correction, b) the functional maps after sign correction, c) the projection matrix after sign correction. Each entry is colored according to its shape class. Note the clusters that form for the functional maps and projection matrices after sign correction.\vspace{0.4cm}}
\label{fig:pca_plot}
}]

\appendix

\noindent This supplementary material presents several components of our work that were not included in the main text:
\begin{itemize}
    \item Evaluation of the sign correction network (Sec.~\ref{sec:supp:sign_corr})
    \item Additional implementation details (Sec.~\ref{sec:supp:impl_details})
    \item A more thorough comparison with the baselines, ablation studies, and qualitative examples (Sec.~\ref{sec:supp:add_eval})
    \item Pseudocode of all algorithms (Sec.~\ref{sec:supp:algorithms})
\end{itemize}

\section{Sign Correction}
\label{sec:supp:sign_corr}

This section provides a quantitative and qualitative evaluation of the sign correction network.
This component of our model allows to select a specific sign for eigenvectors of the Laplacian by projecting them onto the learned feature vectors and making the projection positive (Sec.~\ref{ssec:sign_correction_part}).
After performing the sign correction on all eigenvectors, we obtain a specific basis, independent of the initial one provided by the numerical eigensolver.

\subsection{Accuracy}
\label{sec:sub:results_sign_corr}

\begin{table}[t]
    \small
    \centering
    \begin{tabular}{ccccccc}
    \hline \textbf{\# Ev} & \textbf{F\_r} & \textbf{S\_r} & \textbf{S'19\_r} & \textbf{F\_a} & \textbf{S\_a} \\
    \hline
    \textbf{32} & 98.8 & 99.3 & 98.9 & 98.7 & 99.0 \\
    \textbf{64} & 98.7 & 97.9 & 98.1 & 98.2 & 97.8 \\
    \textbf{96} & 98.7 & 94.9 & 96.3 & 95.1 & 94.7 \\
    \hline
    \end{tabular}
    \caption{Mean Sign Correction Accuracy in $\%$ for each evaluation dataset, averaged over 100 epochs.}
    \label{tab:sign_error}
\end{table}

We first quantitatively evaluate the sign correction network on the human datasets considered in Sec.~\ref{ssec:nearisometric}-\ref{ssec:anisotropic}. 
We use a metric that we call the Mean Sign Correction Accuracy: For each mesh, we calculate the 32, 64, and 96-dimensional eigenbasis twice, perform the sign correction (Eq.~\ref{eq:sign_correction}), and report the mean number of equal eigenvectors.
The metric is averaged over 100 epochs for each dataset.
The results are shown in Table \ref{tab:sign_error}: We observe that the trained sign corrector achieves a high accuracy of ${>}95\%$ on all datasets.

\begin{figure*}[t]
    \centering
    \vspace{1cm}
    \begin{subfigure}[t]{0.33\linewidth}
        \centering
        \includegraphics[width=0.85\linewidth]{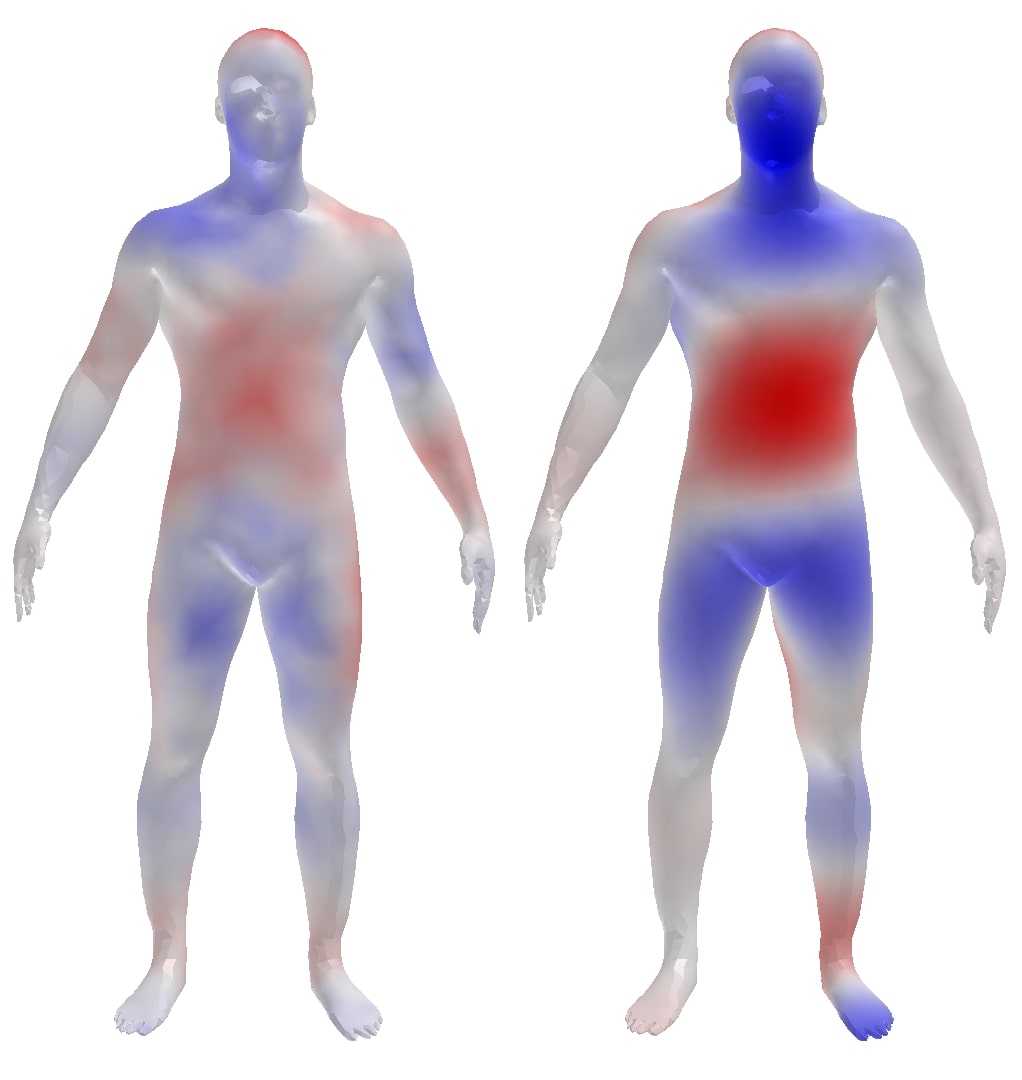}
        \caption{Learned feature vector $\varsigma_{24}$ (left) and the corresponding eigenvector $\phi_{24}$ (right). The projection is $0.51$, so we keep the sign of the eigenvector.}
    \end{subfigure}
    \hspace{1cm}
    \begin{subfigure}[t]{0.33\linewidth}
        \centering
        \includegraphics[width=0.85\linewidth]{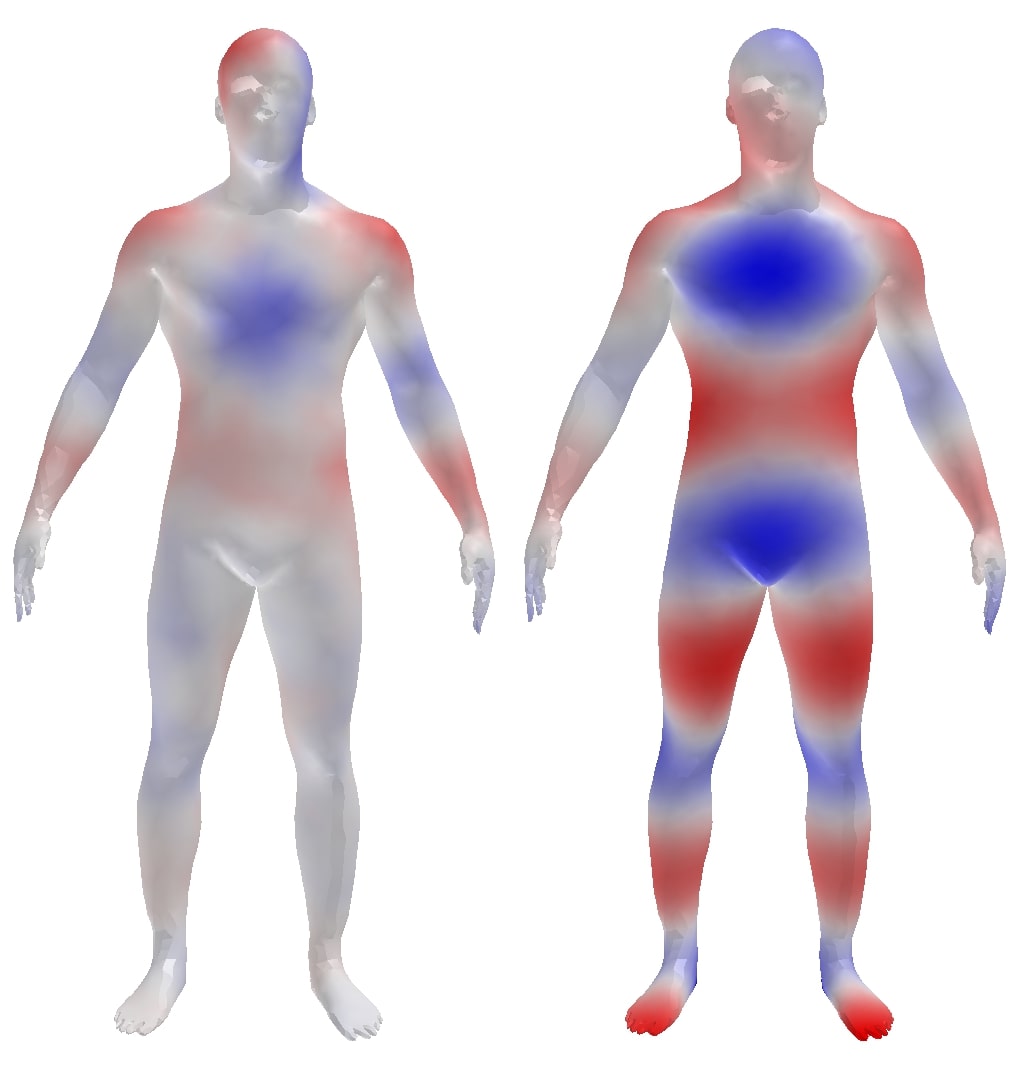}
        \caption{Same for $\varsigma_{32}$ and $\phi_{32}$. The projection is $0.24$.}
    \end{subfigure}
    
    \begin{subfigure}[t]{0.33\linewidth}
        \centering
        \includegraphics[width=0.85\linewidth]{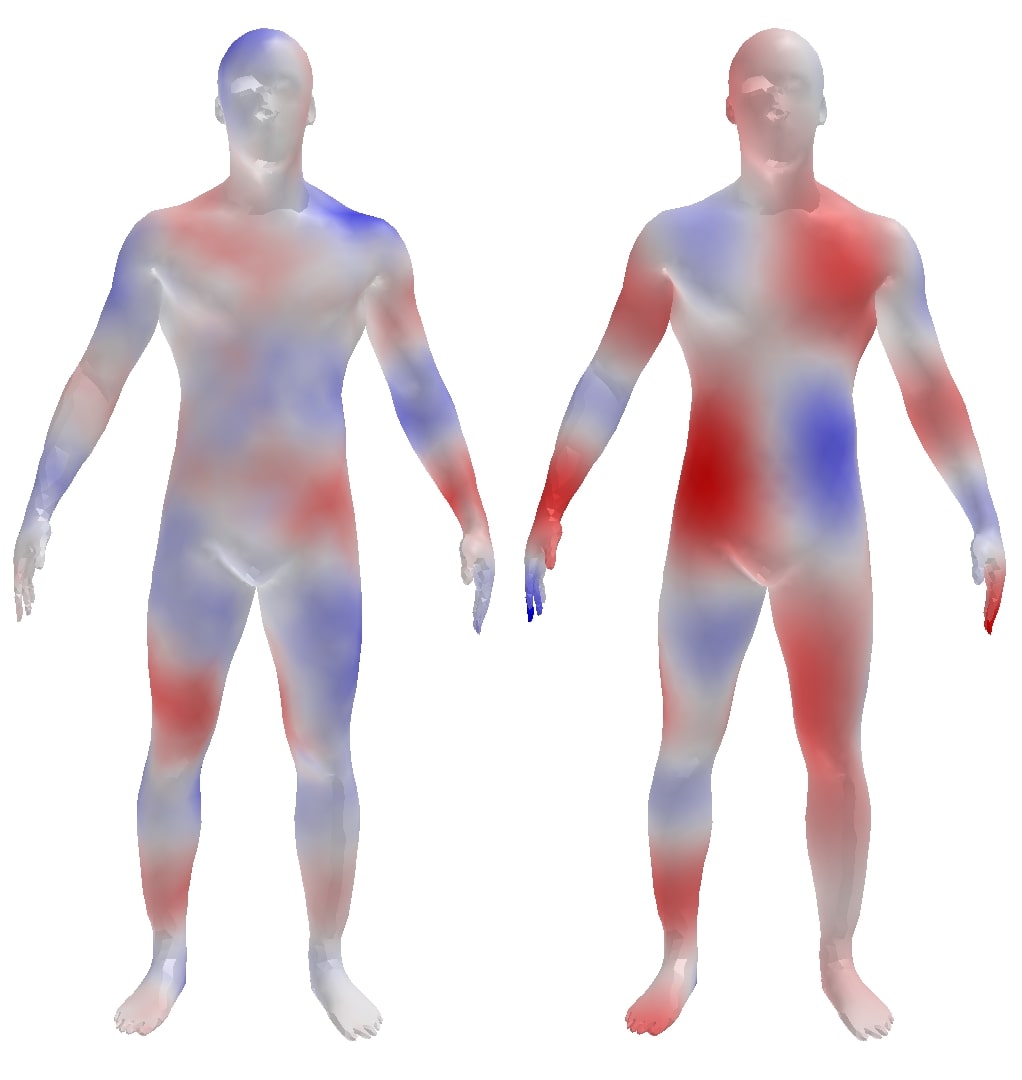}
        \caption{Same for $\varsigma_{40}$ and $\phi_{40}$. The projection is $-0.18$, so the eigenvector should be multiplied by $-1$.}
    \end{subfigure}
    \hspace{1cm}
    \begin{subfigure}[t]{0.33\linewidth}
        \centering
        \includegraphics[width=0.85\linewidth]{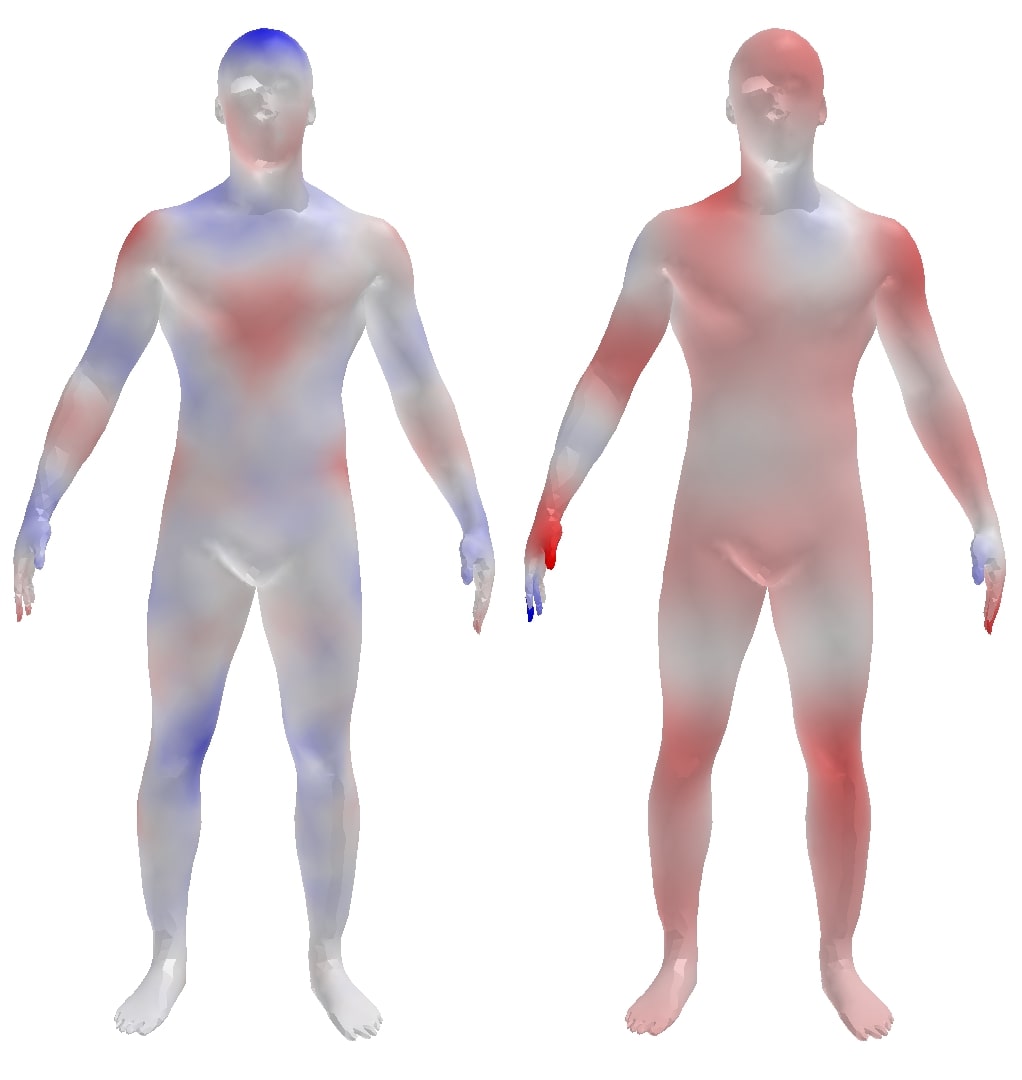}
        \caption{Same for $\varsigma_{48}$ and $\phi_{48}$. The projection is $-0.35$, so the eigenvector should be multiplied by $-1$.}
    \end{subfigure}
    
    \begin{subfigure}[t]{0.33\linewidth}
        \centering
        \includegraphics[width=0.85\linewidth]{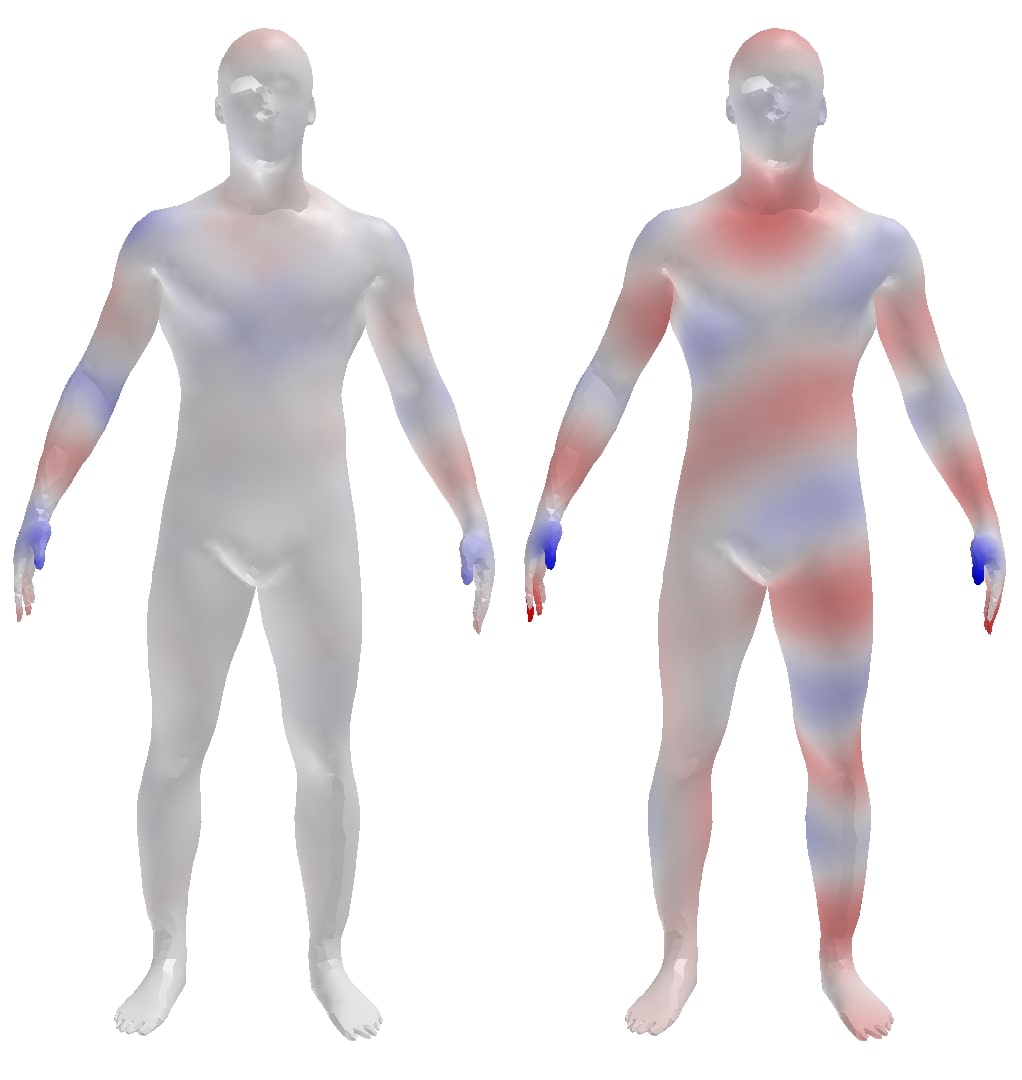}
        \caption{Same for $\varsigma_{56}$ and $\phi_{56}$. The projection is $0.69$.}
    \end{subfigure}
    \hspace{1cm}
    \begin{subfigure}[t]{0.33\linewidth}
        \centering
        \includegraphics[width=0.85\linewidth]{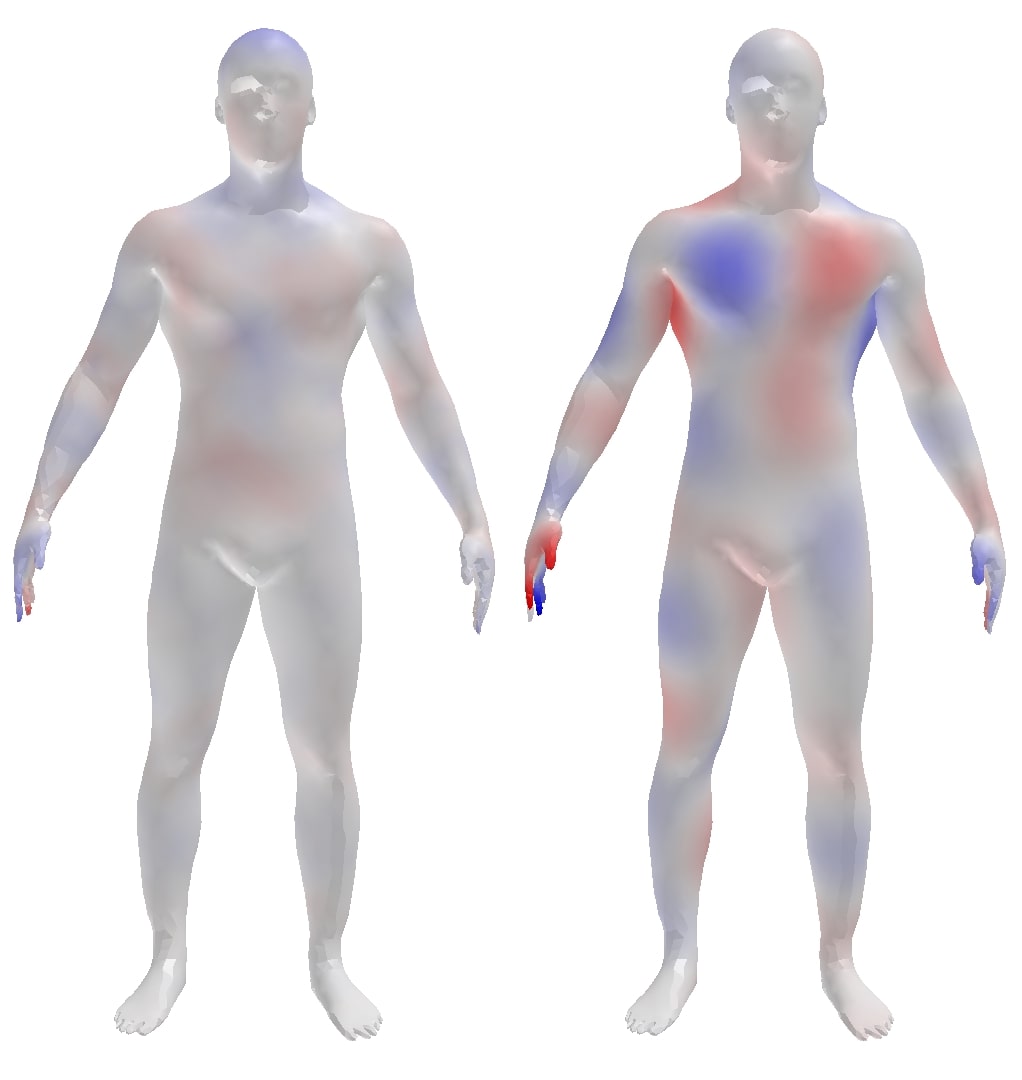}
        \caption{Same for $\varsigma_{64}$ and $\phi_{64}$. The projection is $-0.29$, so the eigenvector should be multiplied by $-1$. Note the weights on the left hand (from our point of view).}
    \end{subfigure}

    \vspace{0.5cm}
    \caption{Learned feature vectors and their corresponding eigenvectors. Positive and negative values are shown in red and blue, respectively. The low-order feature vectors resemble the eigenvectors themselves, while the high-order ones are mainly concentrated in the arm and hand regions.}
    \label{fig:summary_vector_evecs}
\end{figure*}

\subsection{Distribution of Functional Maps}
 
Here we study the change in the distribution of functional maps after sign correction.
We use the FAUST \cite{bogo2014faust} dataset, which contains meshes of 10 classes of humans with various body types in different poses.
For each mesh, we obtain a 32-dimensional eigenbasis and correct it with a pretrained sign correction network.
We then compute the template-wise functional map and the conditioning matrix $y$ (Eq.~\ref{eq:conditioning}).

Next, we perform PCA decomposition \cite{abdi2010principal} on the functional maps before and after sign correction, as well as on the conditioning matrix. 
The pairwise plots of the first two components are shown in Fig.~\ref{fig:pca_plot}, where each point is colored according to its class. 
As we can see, before sign correction, the functional maps did not have a sensible distribution since each of them was defined in a random basis.
After resolving the sign ambiguity, the functional maps form noticeable clusters grouped by class. 
For the conditioning matrix, the overall structure resembles the distribution of functional maps after sign correction. 
This explains the basic principle of our method: the sign correction network transformed the space of functional maps into a learnable distribution by making them refer to a specific basis, and the denoising diffusion model approximated this distribution.

\subsection{Learned Features}

We visualize several of the learned feature vectors in Fig.~\ref{fig:summary_vector_evecs}, along with the corresponding eigenvectors.
As we can see, the low-order feature vectors resemble the eigenvectors themselves, while the high-order ones are mostly concentrated in the arm and hand regions.

\section{Additional Implementation Details}
\label{sec:supp:impl_details}

\subsection{Robustness to Basis Ambiguity}
\label{sec:sub:perm_ambiguity}

To make the sign correction network applicable to high-order eigenvectors, we need to make it robust to possible basis ambiguity that arises when an eigenvalue has high multiplicity (see Sec.~\ref{sec:sub:sign_ambiguity}). 
Numerically, an eigenvalue with multiplicity $d > 1$ is represented as $d$ adjacent eigenvalues $\lambda_i  \dots \lambda_{i+d}$ with close values. 

A straightforward way to account for the basis ambiguity is to use a single feature vector $\varsigma_i$ for several adjacent eigenvectors $\left(\phi_i, \phi_{i+1}\ldots \right)$ instead of only one eigenvector $\phi_i$. 
In practice, we split the eigenvectors into groups of $32$ and select increasingly more eigenvectors per feature vector for each group: one for $\phi_1 - \phi_{32}$, two for $\phi_{33} - \phi_{64}$, and four for $\phi_{65} - \phi_{96}$. 
This takes into account the fact that high-order eigenvalues are more likely to have high multiplicity.

\subsection{Training the Sign Corrector}

The training process described in Sec.~\ref{ssec:training_sign_corr} is based on correcting the signs of two sets of eigenvectors $\Phi_1$ and $\Phi_2$ on the mesh $S$, which can be obtained by performing eigendecomposition with a numerical solver twice.
Due to the sign ambiguity, the difference between them can be written as $\Phi_2 = \Phi_1 \, \sigma$, where $\sigma \in \{-1, 1\}^n$ is the ground-truth sign difference. 
However, we do not need to perform the costly eigendecomposition at each training iteration.
Instead, we can compute the eigenbasis $\Phi$ once, and during training randomly sample two sign combinations $\sigma_1, \sigma_2 \in \{-1, 1\}^n$ to obtain new basis combinations $\Phi_1 = \Phi \sigma_1, \Phi_2 = \Phi \sigma_2$.
This significantly reduces the training time.

\begin{figure}[t]
    \centering
    \includegraphics[width=0.8\linewidth]{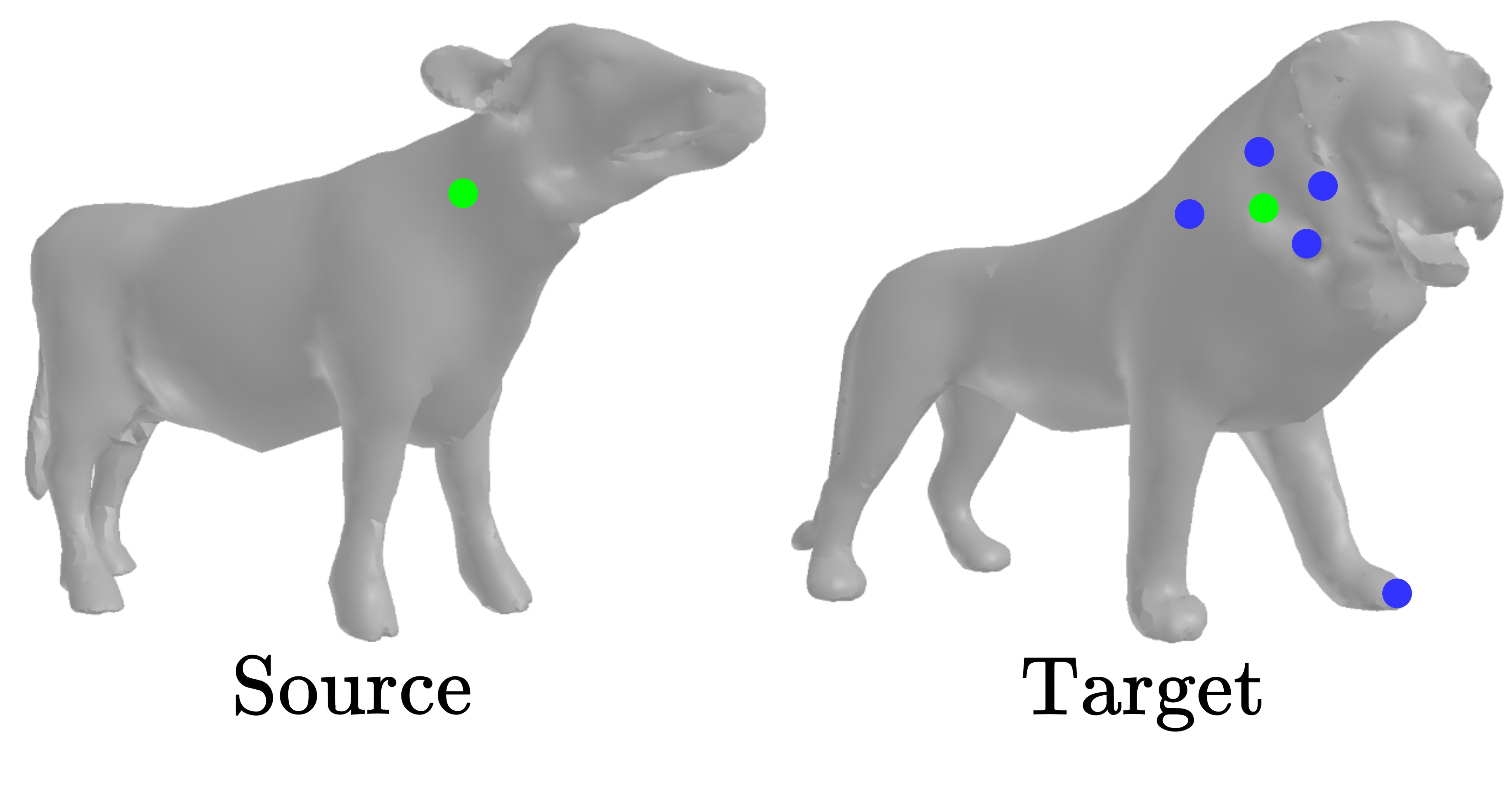}
    \caption{Illustration of the Dirichlet Medoid map selection. Given $k=6$ candidate matches on the target shape, we report the one that has the lowest total distance to the other $k-1$ candidates (shown in green).}
    \label{fig:dirichlet_median}
\end{figure}

\subsection{Map Selection}
\label{ssec:map_selection_supp}

Here, we provide more details about our map selection step (Sec.~\ref{sec:sub:dirichlet_selection}).
We repeat the denoising process $n$ times and rank the candidate maps based on their Dirichlet energy (i.e.~map smoothness).
If we simply report the map with the lowest Dirichlet energy, it may be prone to outliers such as smooth maps with flipped left-right symmetry.
To address this, we select $k$ maps with the lowest Dirichlet energy and iterate over each point on the source shape to combine them. 
At each step, we have $k$ candidate matching points on the target shape, potentially containing outliers.

We find the \textit{medoid} match (the one that has the lowest total distance to the other $k-1$ candidates) and report it; see Fig.~\ref{fig:dirichlet_median} for an illustration.
After repeating these steps for each point on the source shape, we construct a map that we call the ``Dirichlet Medoid''.
In practice, we sample a total of $n=128$ raw maps and use $k=16$ for the medoid selection.

\section{Additional Evaluation}
\label{sec:supp:add_eval}
\begin{table}[t]
    \small
    \centering
    \begin{tabular}{l|ccc}
    \hline 
    \textbf{Test} & \textbf{F\_r} & \textbf{S\_r} & \textbf{S'19\_r} \\
    \hline
    GeomFMaps & 1.9 & 2.4 & 7.5  \\
    ConsistentFMaps & 2.2 & 2.3 & 4.3  \\
    DiffZO & 1.9 & \secondplace{2.2} & 3.6 \\
    ULRSSM & \textbf{1.6} & \textbf{2.1} & 4.6 \\
    SSL & 2.0 & 3.1 & 4.0 \\
    SmS (with align.) & 2.7 & 2.5 & \textbf{3.2} \\
    SmS (no align.) & 3.4 & 3.2 & 4.0 \\    
    \hline 
    \textbf{Ours} -- 64$\times$64 & 1.8 & 2.3 & \secondplace{3.5} \\
    \textbf{Ours} -- 96$\times$96 & \secondplace{1.7} & \textbf{2.1} & 3.9 \\  
    \hline
    \end{tabular}
    \caption{Comparison with descriptor-based methods trained on FAUST+SCAPE. The \textbf{best} and \secondplace{2nd best} results are highlighted. SmS \cite{cao2024spectral} requires rigid pre-alignment of training and test shapes to the same orientation, so we evaluated it on both aligned and unaligned datasets. Note that our method is fully intrinsic and insensitive to rigid orientation.
}
    \label{tab:near_isometric_fs}
\end{table}

\subsection{Qualitative Examples}
\label{sec:qualitative_supp}

\begin{figure*}[t]
    \centering
    \vspace{1cm}
    \begin{subfigure}[t]{\linewidth}
        \centering
        \includegraphics[width=\linewidth]{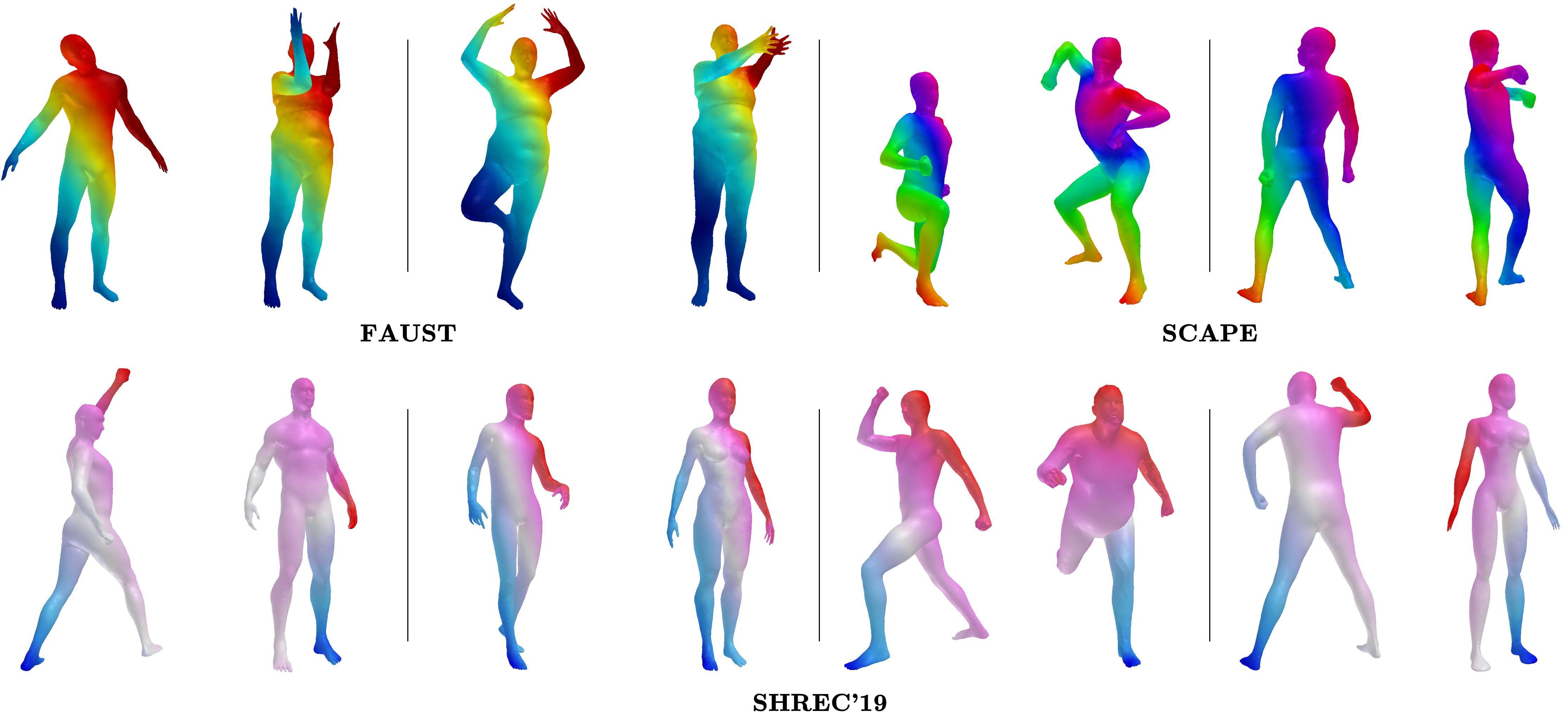}
        \vspace{0.1cm}
        \caption{Examples on human shapes from the FAUST, SCAPE, and SHREC'19 datasets.}
        \label{fig:qualitative_supp_fssh}
        
    \end{subfigure}

    \begin{subfigure}[t]{\linewidth}
        \centering
        \vspace{1cm}
        \includegraphics[width=\linewidth]{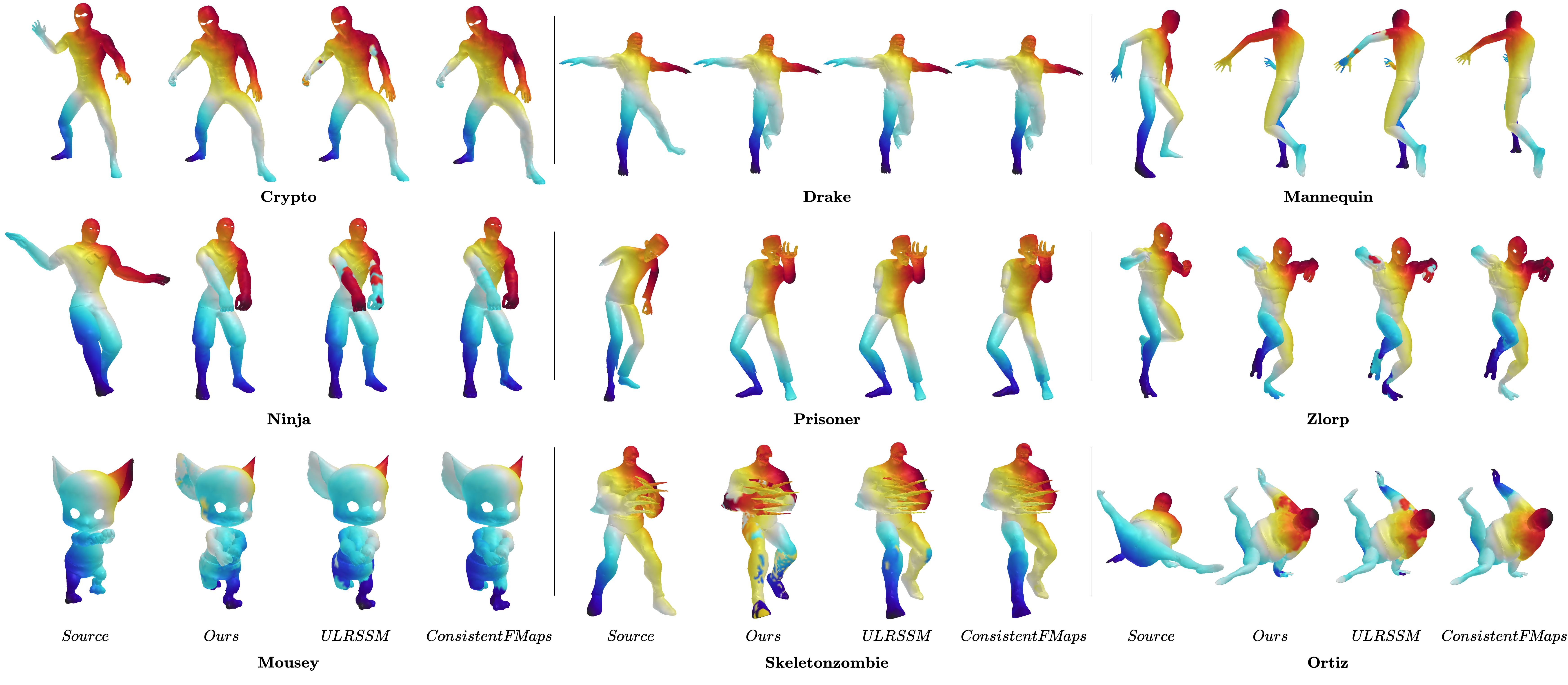}
        \vspace{0.1cm}
        \caption{Examples on the DT4D intra-category dataset, compared to ULRSSM \cite{cao2023Unsupervised} and ConsistentFMaps \cite{sun2023spatially}, state-of-the-art descriptor-based approaches. Our method outperforms them on most shape classes, except for a few categories that are too different from the training data: ``Mousey'', ``Skeletonzombie'', ``Ortiz''. The correspondences for the ``Drake" class are mostly correct but sometimes have flipped left-right symmetry.}
        \label{fig:qualitative_supp_dt4d}
    \end{subfigure}
    \vspace{0.5cm}
    \caption{Additional qualitative examples. Our method provides correct correspondences for challenging human and humanoid meshes.} 
\end{figure*}

Fig.~\ref{fig:qualitative_supp_fssh} shows qualitative examples of correspondences obtained by our method on FAUST, SCAPE, and SHREC'19 datasets.
Fig.~\ref{fig:qualitative_supp_dt4d} qualitatively compares our model with two baselines on all shape classes from the DT4D dataset.
Our method obtains smooth and accurate correspondences for human-like shapes with diverse body types and poses, but it is not as accurate on classes that are too different from the training data.

\subsection{Baselines Trained on FAUST and SCAPE}
\label{sec:results_baselines_faust_scape}

Previously we evaluated descriptor-based methods that were trained on the FAUST dataset since it contains diverse SMPL-like shapes, while SCAPE is too small to fine-tune large models (Sec.~\ref{ssec:nearisometric}).
For a complete comparison, here we additionally evaluate baselines trained on both FAUST and SCAPE.
The results are shown in Table \ref{tab:near_isometric_fs}.
As we can see, our method still performs \textit{on par} with state-of-the-art methods, even without additional variety provided by the shapes from SCAPE.

\begin{table}[t]
    \centering
    \small
    \begin{tabular}{l|ccc}
    \hline 
    \textbf{Test} & \textbf{F} & \textbf{S} & \textbf{S'19} \\
    \hline
    GeomFMaps & 1.9 & 2.5 $\uparrow$ & 7.1 $\downarrow$ \\
    ULRSSM & 1.6 & 2.5 $\uparrow$ & 4.8 $\uparrow$ \\
    SSL & 2.0 & 3.2 $\uparrow$ & 4.4 $\uparrow$ \\
   
    \hline
    \end{tabular}
    \caption{Results of descriptor-based methods trained on 5,000 shapes from SURREAL. Arrows indicate the change in error compared to FAUST+SCAPE (see Table \ref{tab:near_isometric_fs}).}
    \label{tab:descriptor_surreal}
\end{table}

\subsection{Baselines Trained on SURREAL}

Here, we study the performance of other functional map-based methods with respect to dataset size.
We trained several baselines on 5,000 randomly selected shapes from SURREAL and evaluated them.
As shown in Table \ref{tab:descriptor_surreal}, the large dataset size negatively impacts the performance of previous works, with slightly worse results compared to training on FAUST and SCAPE.

Additionally, we note that DiffusionNet-based methods require storing the full decomposition of the Laplacian, which does not scale well to large datasets.
For the full SURREAL dataset, this would require about 1 TB of space.
Instead, our model only stores the conditioning matrices and the functional maps, which takes at most $24$ GB.

\begin{table}[t]
    \small
    \centering
    \begin{subtable}{\linewidth}
        \centering
        \begin{tabular}{c|cc|cc}
            \hline
            & \multicolumn{2}{c|}{\textbf{64$\times$64}}& \multicolumn{2}{c}{\textbf{96$\times$96}}\\
     & \textbf{no ZO}& \textbf{Ours}&\textbf{no ZO} &\textbf{Ours}\\
            \hline
            \textbf{F} & 2.0  &1.8& 1.8  &1.7\\
            \textbf{S} & 2.6  &2.3& 2.2  &2.1\\
            \textbf{S19} & 4.1  &3.5& 4.1  &3.9\\
            \hline
        \end{tabular}
        \caption{Results on FAUST, SCAPE, and SHREC'19 datasets.}
        \label{tab:nozo_fss19}
    \end{subtable}
    
    \vspace{0.5cm} %
    
    \begin{subtable}{\linewidth}
        \centering
        \begin{tabular}{l|cc}
        \hline
 & \multicolumn{2}{c}{\textbf{32$\times$32}}\\
     &\textbf{no ZO}& \textbf{Ours} \\
    \hline
manneq.  &2.0& 1.0 \\
zlorp  &2.2& 1.1 \\
crypto  &2.3& 1.1 \\
prisoner  &3.9& 1.1 \\
ninja  &2.7& 1.4 \\
ortiz  &19.5& 9.3 \\
mousey  &18.1& 10.2 \\
drake  &12.5& 10.6 \\
skeletonz.  &33.2& 16.3 \\
    \hline
    \end{tabular}
    \caption{Results on the DT4D dataset.}
    \label{tab:nozo_dt4d}
    \end{subtable}
    
    \caption{Results of our models with and without ZoomOut \cite{melzi2019zoomout} refinement.}
\end{table}

\subsection{Impact of ZoomOut}

To illustrate the role that ZoomOut refinement \cite{melzi2019zoomout} plays in our pipeline, we tested our models without the spectral upsampling step. 
The results are shown in Tables \ref{tab:nozo_fss19}-\ref{tab:nozo_dt4d}.
As we can see, ZoomOut plays a key role for 32$\times$32 and 64$\times$64 models, while it is less important for the 96$\times$96 one.

\subsection{DT4D: Impact of the Model Dimension}
\label{sec:supp:d4td_basis_number}

Our previous experiments show that models predicting high-dimensional functional maps achieve higher accuracy, while low-dimensional ones provide better generalization to unseen shape classes.
To illustrate this, we show in Table \ref{tab:d4td_basis_number} the results of all our models on shape classes from DT4D. 
As we can see, the 32$\times$32 model achieves excellent performance on most shape classes, while the 96$\times$96 one shows high accuracy on a few specific classes but fails on others.
The 64$\times$64 model falls in between.

\begin{table}[t]
    \small
    \centering
    \begin{tabular}{l|ccc|ccc}
    \hline
    \multirow{2}{*}{\textbf{Class}} & \multicolumn{3}{c|}{\textbf{Intra}} & \multicolumn{3}{c}{\textbf{Inter}} \\
    & \textbf{32}& \textbf{64}& \textbf{96}& \textbf{32}& \textbf{64}& \textbf{96}\\
    \hline
manneq. & 1.0& 1.0&1.0& 3.3& 3.2& 3.2\\
zlorp & 1.1& 1.1&1.1& 4.2& \textbf{3.6}& 4.4\\
crypto & 1.1& 1.1&\textbf{1.0}& -- & -- & --\\
prisoner & \textbf{1.1} & 2.3&9.1& 29.6 & 28.1& 34.5\\
ninja & \textbf{1.4} & 2.0&9.3& \textbf{4.3} & 5.2& 9.5\\
ortiz & 9.3 & 17.0&24.2& -- & -- & -- \\
mousey & 10.2 & 17.9&25.6& -- & -- & -- \\
drake & 10.6 & \textbf{6.8}&9.4& 10.4 & 8.5& 13.2\\
skeletonz. & 16.3 & 30.1&36.5& 49.4 & 31.5& 39.3\\
    \hline
    \end{tabular}
    \caption{Results of our 32$\times$32, 64$\times$64, and 96$\times$96 models on shape classes from DT4D-H.}
    \label{tab:d4td_basis_number}
\end{table}

\subsection{Ablation Study}
\label{ssec:ablations} 

\begin{table}[t]
    \small
    \centering
    \begin{tabular}{lc}
        \hline Ablation Setting & SHREC'19 \\
        \hline
        w/o basis ambiguity (Sec. \ref{sec:sub:perm_ambiguity}) & 5.8 \\
        w/o ZoomOut & 4.1 \\
        w/o any map selection (Sec. \ref{sec:sub:dirichlet_selection}) & 4.7 \\
        w/o medoid map selection (Sec. \ref{ssec:map_selection_supp}) & 3.7 \\
        Ours & \textbf{3.5} \\
        \hline
    \end{tabular}
    \caption{Ablation study of the 64$\times$64 model on SHREC'19.}
    \label{tab:ablation}
\end{table}

\begin{figure*}[t]
    \centering
    \begin{subfigure}[t]{0.3\linewidth}
        \centering
        \includegraphics[width=\linewidth]{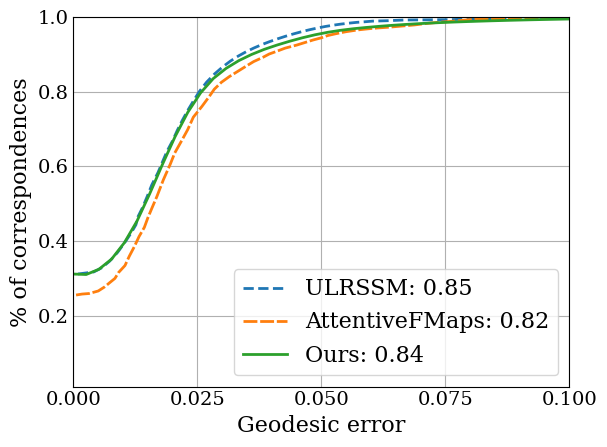}
        \caption{FAUST}
    \end{subfigure}
    \begin{subfigure}[t]{0.3\linewidth}
        \centering
        \includegraphics[width=\linewidth]{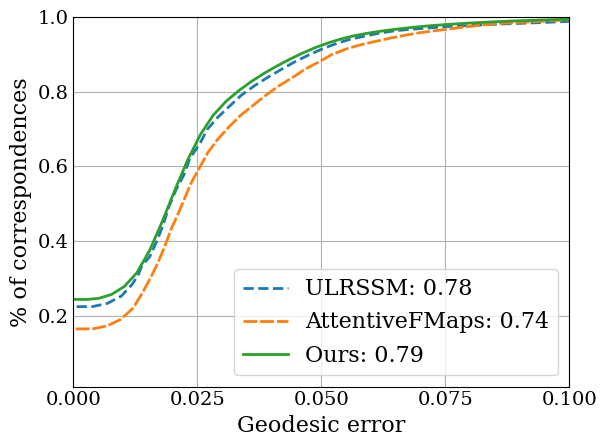}
        \caption{SCAPE}
    \end{subfigure}
    \begin{subfigure}[t]{0.3\linewidth}
        \centering
        \includegraphics[width=\linewidth]{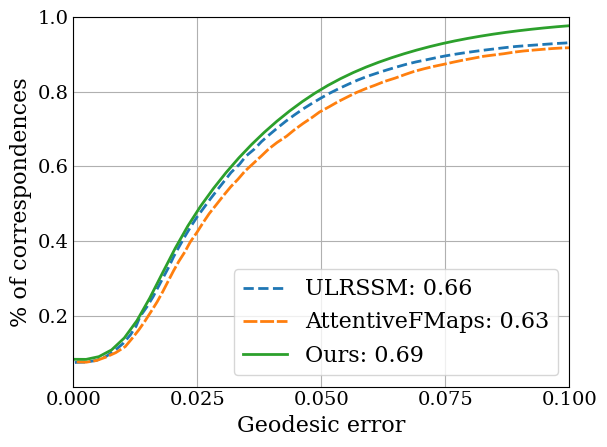}
        \caption{SHREC'19}
    \end{subfigure}
    \caption{PCK curves for the evaluation datasets. Our method demonstrates superior or similar performance compared to existing approaches.}
    \label{fig:pck_curves}
\end{figure*}

\begin{figure*}[t]
    \centering
    \includegraphics[width=0.9\linewidth]{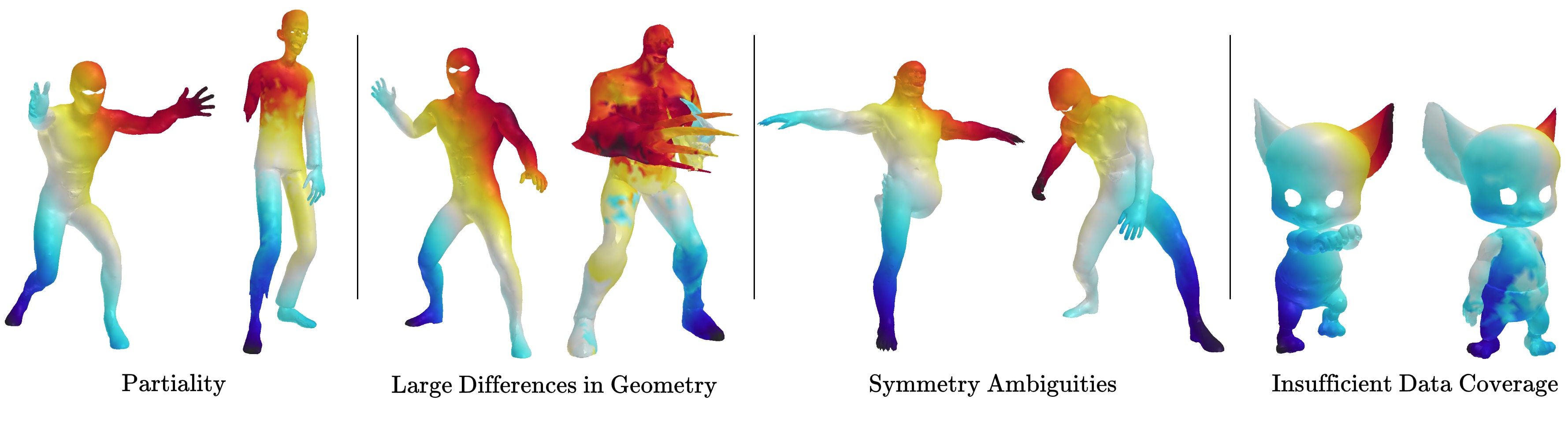}
    \caption{Limitations of our method. They include shapes with large differences from the synthetic human data, as well as topological inconsistencies like missing body parts. These limitations can be addressed in future work.}
    \label{fig:limitations}
\end{figure*}

Next, we ablate different parts of our method. 
We use a 64$\times$64 model and the SHREC'19 dataset since it provides the largest shape diversity; see Table \ref{tab:ablation} for the results. 
As we can see, addressing the basis ambiguity (Sec. \ref{sec:sub:perm_ambiguity}) is crucial for the matching performance. 
Second, using ZoomOut \cite{melzi2019zoomout} to increase the resolution of the functional map helps to improve the accuracy. 
Finally, using map selection criteria based on Dirichlet energy is also a necessary component of our model, and selecting the medoid among several smoothest maps (rather than using the map with the lowest Dirichlet Energy) helps to filter out the outliers.

\subsection{Comparison with ULRSSM on DT4D}
\label{sec:ulrssm_dt4d_supp}

In Sec.~\ref{ssec:zeroshot}, we evaluated our method on shape classes from the DT4D dataset against the baselines, ConsistentFMaps \cite{sun2023spatially} and ULRSSM \cite{cao2023Unsupervised}.
Compared to other descriptor-based methods, ULRSSM fine-tunes the parameters of the feature extractor by backpropagating the unsupervised loss during test time for each evaluation pair.
For a complete comparison, in Table~\ref{tab:d4td_ulrssm_noref} we evaluate ULRSSM without refinement, which leads to worse results of this baseline. 

\begin{table}[t]
    \small
    \centering
    \small
    \begin{tabular}{l|ccc|ccc}
    \hline
    \multirow{3}{*}{\textbf{Class}} & \multicolumn{3}{c|}{\textbf{Intra}} & \multicolumn{3}{c}{\textbf{Inter}} \\
    \cline{2-7}
    & \multirow{2}{*}{\textbf{Ours}} & \multicolumn{2}{c|}{\cite{cao2023Unsupervised}}& \multirow{2}{*}{\textbf{Ours}} & \multicolumn{2}{c}{\cite{cao2023Unsupervised}}\\
 & & \textbf{FT}& \textbf{no FT}& & \textbf{FT}&\textbf{no FT}\\
    \hline
manneq. & 1.0 & 1.0&3.4& \textbf{3.3} & 4.0& 11.3\\
zlorp & \textbf{1.1} & 1.3&5.2& \textbf{4.2} & 6.9 & 21.3\\
crypto & \textbf{1.1} & 1.5&3.3& -- & -- & --\\
prisoner & 1.1 & 1.1&1.4& 29.6 & 20.5 & 33.0\\
ninja & \textbf{1.4} & 5.6&11.7& \textbf{4.3} & 16.2 & 30.6\\
ortiz & 9.3 & \textbf{4.4}&7.8& -- & -- & -- \\
mousey & 10.2 & \textbf{2.7}&5.2& -- & -- & -- \\
drake & 10.6 & \textbf{1.0}&2.6& 10.4 & 8.2  & 26.4\\
skeletonz. & 16.3 & \textbf{1.4}&4.7& 49.4 & 36.5 & 55.1\\
    \hline
    \end{tabular}
    \caption{
    Results of our method on shape classes from DT4D \cite{magnet2022smooth} compared to ULRSSM \cite{cao2023Unsupervised} with and without fine-tuning.
    }
    \label{tab:d4td_ulrssm_noref}
\end{table}

\subsection{PCK Curves}

We show PCK curves for the FAUST, SCAPE, and SHREC'19 datasets in Fig.~\ref{fig:pck_curves}, which plot the percentage of correctly predicted keypoints within different distance thresholds from the ground truth.
Our method demonstrates competitive performance compared to ULRSSM \cite{cao2023Unsupervised} and AttentiveFMaps \cite{li2022learning}.

\subsection{Limitations} 

In Fig.~\ref{fig:limitations}, we illustrate the main limitations of our method.
These include shapes with significant differences from the synthetic human data we used for training, as well as shapes with missing body parts. 
In these cases, our method tends to output maps with incorrect left-right symmetry.
These limitations may be addressed in future work.

\section{Algorithms}
\label{sec:supp:algorithms}

We additionally provide the algorithms discussed in Sec.~\ref{sec:method} as pseudocode.
The training and inference of the diffusion model are described in Alg.~\ref{alg:ddpm_training} and Alg.~\ref{alg:ddpm_inference}, respectively.
The sign correction is described in Alg.~\ref{alg:sign_corr_mass_norm} and the training process in Alg.~\ref{alg:sign_corr_training}.

\clearpage

\begin{algorithm}[h]
\caption{DDPM Training Pipeline}
\label{alg:ddpm_training}

Train the Sign Correction Network $\Theta$\;

Correct all eigenvectors $\Phi$ with $\Theta$ in the dataset\;

Get functional maps to the template, $C_{1T}$\;

Calculate conditioning $y$\;

Train a DDPM to predict $C_{1T}$ given $y$\;

\end{algorithm}

\SetKwBlock{Repeat}{repeat}{}

\begin{algorithm}[h]

\caption{DDPM Inference with Selection}
\label{alg:ddpm_inference}

\KwIn{Shape collection $\{S_i\}$, List of Test Pairs $\mathcal{L}$, Sign Correction Net $\Theta$, DDPM}

\BlankLine
\tcp{Template Stage}

\For{shape $S$, eigenvectors $\Phi$ in $\{S_i\}$}{

\Repeat(\textit{n = 128 times} \tcp*[f]{See \ref{sec:sub:dirichlet_selection}}){

Correct the signs of $\Phi$, get conditioning $y$\;

$C_{1T} \coloneq \textbf{DDPM}\left(y\right)$\;

Convert $C_{1T}$ to pointwise map $\Pi_{T1}$\;
}
}

\BlankLine

\tcp{Pairwise Stage}

\For{shapes $S_1, S_2$ in $\mathcal{L}$}{

$\mathcal{T}_\Pi \coloneq \{\}$\;

\BlankLine
\Repeat(\textit{n = 128 times} \tcp*[f]{See \ref{sec:sub:dirichlet_selection}}){
$C_{12} \coloneq \textbf{LstSq}\left( \Pi_{T2} \Phi_2, \Pi_{T1} \Phi_1 \right)$\;

Zoomout $C_{12}$ to $[200, 200]$\;

Convert $C_{12}$ to pointwise map $\Pi_{21}$\;

Add $\Pi_{21}$ to $\mathcal{T}_{\Pi}$\;
}

\BlankLine

$\hat{\Pi}_{21} \coloneq \textbf{Selection}(\mathcal{T}_\Pi)$ 
}

\KwOut{pointwise maps $\hat{\Pi}_{21}$ for each pair}
\end{algorithm}

\newpage

\begin{algorithm}[h]
\caption{Learned Sign Correction}
\label{alg:sign_corr_mass_norm}
\KwIn{Shape $S$, eigenvectors $\Phi \sim $ (v, n), vertex-area matrix $A \sim$ (v, v), feature extractor $\Theta$}

\BlankLine

$\Sigma \coloneq \, \Theta\left(S\right)$
\tcp*{See \ref{sec:sub:perm_ambiguity}, $(v,n)$}

\BlankLine

$P \coloneq \Sigma^T A \Phi$
\tcp*{$(n,n)$}

\eIf{
training
}{
$\hat{\sigma} \coloneq \textbf{Diag}(P)$
\tcp*{$(n)$}
}{
$\hat{\sigma} \coloneq \textbf{Sign}\left(\textbf{Diag}(P)\right)$
\tcp*{$(n)$}
}

$\hat{\Phi} \coloneq \Phi \hat{\sigma}$
\tcp*{$(v,n)$}

\BlankLine

\KwOut{Eigenvectors with corrected signs $\hat{\Phi}$,
correcting sign sequence $\hat{\sigma}$}

\end{algorithm}

\begin{algorithm}[h]
\caption{Unsupervised Training of Sign Correction}
\label{alg:sign_corr_training}
\KwIn{Shape collection $\{S_i\}$, feature extractor $\Theta$}

\BlankLine

\For{N epochs}{
    \For{shape $S$, eigenvectors $\Phi$ in $\{S_i\}$}{

        \BlankLine
        \tcp{$-1$ or $1$}
        $\sigma_1 \coloneq \textbf{RandomSigns}\left(n\right)$\;
        $\sigma_2 \coloneq \textbf{RandomSigns}\left(n\right)$\;
        $\Phi_1 \coloneq \Phi \sigma_1$\;
        $\Phi_2 \coloneq \Phi \sigma_2$\;

        \BlankLine
        \tcp{Alg.\ref{alg:sign_corr_mass_norm}}
        $\hat{\sigma}_1 \coloneq \textbf{SignCorr}\left(S, \Phi_1, \Theta\right)$\;
        $\hat{\sigma}_2 \coloneq \textbf{SignCorr}\left(S, \Phi_2, \Theta\right)$\;

        \BlankLine
        $\mathcal{L}_{\operatorname{sign}} \coloneq \textbf{MSE}(\sigma_1 \sigma_2, \hat{\sigma}_1 \hat{\sigma}_2)$\;
        \textbf{Backpropagate}$(\mathcal{L}_{\operatorname{sign}})$
    
    }
}
\KwOut{Trained feature extractor $\hat{\Theta}$}

\end{algorithm}

\end{document}